\documentclass{article}

\usepackage{PRIMEarxiv}

\usepackage[utf8]{inputenc} 
\usepackage[T1]{fontenc}    
\usepackage{hyperref}       
\usepackage{url}            
\usepackage{booktabs}       
\usepackage{amsfonts}       
\usepackage{nicefrac}       
\usepackage{microtype}      
\usepackage{lipsum}
\usepackage{fancyhdr}       
\usepackage{graphicx}       
\graphicspath{{media/}}     
\usepackage{amsmath}
\usepackage{subcaption}
\usepackage{caption}

\usepackage{makecell}
\usepackage{rotating}
\usepackage[colorinlistoftodos,prependcaption,textsize=small]{todonotes}

\usepackage[backend=bibtex, style=numeric, defernumbers]{biblatex}

\title{Modeling Events and Interactions through Temporal Processes - A Survey}

\author{
    Angelica Liguori \\
    University of Calabria and \\ Institute for High Performance Computing and Networking, \\ Italian National Research Council (ICAR-CNR) \\
    Italy \\
    \texttt{angelica.liguori@dimes.unical.it} \\
    \And
    Luciano Caroprese \\
    University G. d'Annunzio of Chieti-Pescara \\
    Italy \\
    \texttt{luciano.caroprese@unich.it} \\
    \And
    Marco Minici \\
    University of Pisa and \\ Institute for High Performance Computing and Networking, \\ Italian National Research Council (ICAR-CNR) \\
    Italy \\
    \texttt{marco.minici@icar.cnr.it} \\
    \And
    Bruno Veloso \\
    Faculty of Economics - University of Porto and \\ Institute for Systems and Computer Engineering, \\ Technology and Science (INESC TEC) \\
    Portugal \\
    \texttt{bveloso@fep.up.pt} \\
    \And
    Francesco Spinnato \\
    Scuola Normale Superiore of Pisa and \\ Institute of Information Science and Technologies, \\ Italian National Research Council (ISTI-CNR) \\
    Italy \\
    \texttt{francesco.spinnato@sns.it} \\
    \And
    Mirco Nanni \\
    Institute of Information Science and Technologies,\\ Italian National Research Council (ISTI-CNR) \\ and KDDLab \\
    Italy \\
    \texttt{mirco.nanni@isti.cnr.it} \\
    \And 
    Giuseppe Manco \\
    Institute for High Performance Computing and Networking, \\ Italian National Research Council (ICAR-CNR) \\
    Italy \\
    \texttt{giuseppe.manco@icar.cnr.it} \\
    \And
    Jo\~ao Gama \\
    Faculty of Economics - University of Porto and \\ Institute for Systems and Computer Engineering, \\ Technology and Science (INESC TEC) \\
    Portugal \\
    \texttt{jgama@fep.up.pt}
}

\begin{document}

\maketitle

\begin{abstract}
In real-world scenario, many phenomena produce a collection of events that occur in continuous time. Point Processes provide a natural mathematical framework for modeling these sequences of events. In this survey, we investigate probabilistic models for modeling event sequences through temporal processes. We revise the notion of event modeling and provide the mathematical foundations that characterize the literature on the topic. We define an ontology to categorize the existing approaches in terms of three families: simple, marked, and spatio-temporal point processes. For each family, we systematically review the existing approaches based based on deep learning. Finally, we analyze the scenarios where the proposed techniques can be used for addressing prediction and modeling aspects.
\end{abstract}


\section{Introduction} \label{sec:introduction}

\subsection{Context}

Event modeling refers to the capability of understanding the dynamics of complex processes and characterizing their temporal distribution according to endogenous and exogenous aspects. 
This problem is of scientific and practical relevance since event data is common in many real-world scenarios and sparks interest in many fields including medicine, epidemiology, engineering, earth science, economics, finance, and social science. 

In medicine, events can represent various situations, such as incidents, test results, diagnoses and symptoms, and medications. The advent of wearable devices and apps also allows tracking human activities, such as eating, working, sleeping, traveling, etc. Events also characterize movement patterns such as trajectories or taxi/car/public transportation adoptions. 
In engineering, events can represent phenomena occurring in complex environments, such as failures occurring in industrial processes. In earth science, monitoring and modeling phenomena such as volcano eruptions, seismic events, or floods are of crucial importance. Such phenomena are typically characterized by foregoing events whose interactions can be premonitory of the forthcoming disasters and their aftershocks events. 
Within finance, data relative to sales and purchase events, as well as any other activities that can influence future trades, represents a valuable source of information concerning the market structure, and the analysis of temporal correlations among such events and their stochastic nature is at the basis of valuation and risk management. 
In biology, event modeling is used for characterizing the evolution of populations of individuals. Also, epidemic modeling focuses on contagion events and on characterizing the transmission of a contagious disease within a population. 

More recently, the analysis of social media has focused on the interactions among individuals within content-sharing platforms such as Facebook, Twitter, etc. Here, events can be user actions over time, each of which can be characterized by user properties and content features. Interaction event times represent valuable information for social scientists to gain insight into complex social dynamics and their study can be used to characterize phenomena such as influence, polarization and radicalization, spread of misinformation. 

\subsection{Main Challenges}

There are several aspects that make the problem of modeling events challenging. On one side, beyond temporal occurrences of events, additional information for the entities of interest (such as type of disease/symptom, magnitude and the location of earthquakes, social interactions, etc.) characterize the occurrences of events. 
Given this complex combination of temporal information and additional features (marks), it is necessary to build models able to capture the heterogeneity and complex dependencies among the marks.

Also, uncertainty quantification is a vital component characterizing event modeling. There can be different alternative interpretations of a situation, whose control requires the capability to assimilate information from different sources and to make decisions about what new information would be most useful in order to disambiguate the possible outcomes. However, the complexity of the scenarios under investigation, characterized by an interplay between endogenous and exogenous factors that can change considerably the unfolding of an event sequence, makes it hard to investigate the hidden causal relationships that characterize data and ultimately guarantee robustness and trustability in decisions. 



This is further exacerbated by the intrinsic difficulty of modeling temporal aspects, both in discrete and continuous settings. Time-to-event, i.e., the time elapsed until an event occurs, can only be observed when the event occurs. However, unobserved phenomena are still possible and we have no knowledge about them until they occur. For example, a reaction to a social media post could occur immediately, within two hours or never. This reflects the fact that event types may exhibit different properties and their occurrence can once again be affected dramatically by exogenous causes. In the example above, the lack of a reaction could be due to the fact that the content is not of interest to the user, or to the lack of exposure. The uncertainty coming from a mix of complete and incomplete observations makes probabilistic modeling extremely difficult.

\subsection{Related Surveys}
The above challenges make the underlying research problem intriguing. And in fact, there is an abundance of approaches in the literature, rooted on statistics, machine learning and knowledge discovery. The relevance of the topic is also witnessed by the abundance of surveys covering it, which we summarize below. 

In \cite{paper21}, the authors provide a general overview of neural temporal point processes focusing on general principles and building blocks for constructing neural Temporal Point Process (TPP) models. 
Specifically, the authors focus on models and applications of TPP. The work is focused on auto-regressive neural TPP and continuous-time state evolution models. The authors also provide a section in which they highlight the advantages and disadvantages of the two approaches. For the applications, the authors identify two categories: Prediction tasks (aim to predict the time and/or the type of future events) and structure discovery tasks (learn dependencies between different event types). The authors conclude the survey highlighting the main challenges that the field faces. Compared to this survey, our survey aims to provide an overview considering both neural and non-neural temporal point processes as well as survival approaches.

In \cite{paper22}, the authors highlight how temporal point processes can be used for modeling event sequences in continuous time-space, thus being able to be devised for many possible applications such as online purchases or device failures. The main topic of this work is to provide a literature review that critically analyzes the difference between traditional statistical approaches and more recent neural point processes. 
The authors investigate this problem under the lens of Machine Learning (ML): The duality between model capacity and model interpretability. Traditional TPPs excel at clear interpretability since, via domain knowledge, the intensity function must be explicitly parameterized. On the contrary, neural point processes show a higher capacity to learn unknown distribution.

An overview of (marked) temporal point processes is proposed in \cite{paper23} 
: In particular, the lecture is focused on defining the point processes using the conditional intensity function $\lambda$. The author also provides some literature examples of point processes where the conditional intensity function has functional forms, e.g., Hawkes process, Poisson process. In addition, the author provides a way to estimate the parameters of a point process specified by a conditional, i.e., the maximum likelihood inference. If the condition is specified, the point process can be simulated, and in this lecture two simulation methods are explained.

In \cite{paper51}, the authors provide a review describing statistical methods used to model the spatio-temporal point processes. 
A spatio-temporal point process is a list of events that occurs in a certain time and in a certain space, so the goal is to predict when and where a next event occurs. First, the authors provide a description of the characteristics of those processes and then they focus on the review of the state-of-the-art statistical approaches.
Respect to this survey, our survey aims to provide an overall perspective of the point process considering both statistical and deep learning models and considering not only the spatio-temporal point process but also (un)marked temporal point process as well as (un) marked spatio-temporal point processes.

The paper \cite{paper28} surveys the literature on survival analysis 
, providing: (\textit{i}) A problem definition; (\textit{ii}) an in-depth description of statistical approaches; (\textit{iii}) a brief summary of machine learning approaches (mostly standard ones, very little about deep learning); (\textit{iv}) a few evaluation metrics; (\textit{v}) various application domains; (\textit{vi}) a list of software libraries, virtually all in R.  The main strength of the work is its extent and the detail provided for statistical approaches. The main limitations are in the briefness of ML and Deep Learning (DL) approaches, and the presentation of statistical concepts is sometimes unclear. A taxonomy is given, which is useful, yet a bit “flat” on the ML/DL side. The most recent references (but one) are dated 2017.

In \cite{paper33}, the authors address the problem of Influence Maximization (IM) on Social Graphs 
First, they depict the algorithmic problem by defining the diffusion model, the influence spread function, and the objective function. Then, they focus on the four main typologies of diffusion models - i.e., Linear Threshold (LT), Independent Cascade (IC), Triggering (TR) and the Time-Aware model. After this broad overview of the problem, they prove the NP-hardness of IM. They also show that it is possible to approximate an optimal solution only if the influence spread function satisfies monotonicity and submodular properties. The survey's core is an extensive evaluation of the current algorithmic frameworks. The authors review many works in the literature and identify three categories: Simulation-based, proxy-based, and sketch-based. The first has the advantage of model generality at the cost of computational efficiency to have estimations with small errors. The proxy-based is practically efficient since it can provide polynomial solutions but lacks theoretical guarantees. Sketch-based tries to improve the theoretical efficiency of simulation-based approaches by employing sketches to avoid rerunning Monte Carlo simulations. Finally, the authors analyze emerging trends with a particular emphasis on Context-Aware Influence Maximization.

The book chapter in \cite{paper48} highlights the challenges of exploiting information-cascade data for Social Network Analysis. The main contribution is a general deep learning framework to handle information cascades which comprise six different phases: (\textit{i}) Data collection, (\textit{ii}) data encoding, (\textit{iii}) data representation, (\textit{iv}) deep learning models construction, (\textit{v}) tasking setting, and (\textit{vi}) task solving. Then, for each phase, they highlight possible design choices by describing their modeling capabilities. Finally, from the application point of view, they focus on user behavior analysis, information cascade prediction, rumor detection, and analysis of social network events.
In \cite{paper49}, the authors propose a comprehensive survey of methods for processing the so-called \emph {information cascades}, trajectories, and structures of information diffusion, to predict the adopters/participants in the spreading process and its timing. The analysis range from \emph{feature engineering} and \emph{stochastic processes}, through \emph{graph representation}, to \emph{deep learning-based approaches}. Specifically, the authors first formally define different types of information cascades and summarize the perspectives of existing studies. Then they present a taxonomy that categorizes existing works into the aforementioned three main groups as well as the main subclasses in each group, and they systematically review cutting-edge research work. Finally, they summarize the pros and cons of existing research efforts and outline the open challenges and opportunities in this field.While this survey focuses on the information cascades, our review adopts a broader perspective by analyzing the most recent techniques related to the \emph{Temporal Point Processes} of which information cascades are particular cases.

In \cite{paper52} the existing methodologies developed for event prediction methods in the big data era are presented. It provides an extensive overview of the event prediction challenges, techniques, applications, evaluation procedures, and future outlook, summarizing the research in the last five years. 
Event prediction challenges, opportunities, and formulations have been discussed in terms of the event element to be predicted, including the event location, time, and semantics, after which the authors propose a systematic taxonomy of the existing event prediction techniques according to the formulated problems and types of methodologies designed for the related problems. They also have analyzed the relationships, differences, advantages, and disadvantages of these techniques from various domains, including machine learning, data mining, pattern recognition, natural language processing, information retrieval, statistics, and other computational models. In addition, a comprehensive and hierarchical categorization of popular event prediction applications has been provided that covers domains ranging from natural science to the social sciences. In the paper, the authors also highlight open problems and future trends in this domain.

In \cite{paper78} the authors initially highlight that many human and natural activities are clustered in time and space, thus conveying relevant information about diffusion processes. Then, starting from the previous statement, they argue how current literature ignores universal components of diffusion processes across disciplines, motivating the need for further exploration of this aspect. For this reason, they define a taxonomy of common factors in diffusion dynamics, detecting four different categories: (\textit{i}) Exogenous effects, (\textit{ii}) endogenous effects, (\textit{iii}) diffusion items, and (\textit{iv}) diffusion spaces. Furthermore, they re-analyze and compare modern diffusion models under the theoretical lens of point processes, also relating each of them to the previously defined taxonomy. Finally, as a case study, they propose a general framework for epidemic modeling based on the proposed taxonomy.

Each of the analyzed surveys is focused on a particular aspect. Differently, our survey aims to provide an overall perspective of the point process considering both statistical (non-neural) and deep learning (neural) -based models as well as survival approaches.

\subsection{Contributions}


Given the above discussion, one could ask why we need another survey on the topic. The fact is that the impulse given by the widespread adoption of artificial intelligence techniques in the above mentioned domains, has witnessed a rather fragmented research field. In fact, the traditional statistical approaches developed since the early seventies do not necessarily scale on large volumes of data (characterizing, e.g., social media or large engineering/manufacturing applications). In addition, several key contributions were developed specific to vertical domains (such as medicine or biology) and it is natural to ask whether they can be adapted to more recent scenarios. Finally, the disruptive evolution of machine and deep learning in the recent years, has made several traditional contributions outdated. It is natural hence to look for a systematic comparison and analysis. All these aspects contribute to outline this survey. 

The main contributions of this paper can be summarized as follows:
\begin{itemize}
\item We revisit the notion of event modeling and lay the mathematical foundations that characterize the literature on the topic. 
\item We unveil the unique challenges represented by the modeling of temporal processes in modern real-life, data abundant scenarios.
\item We systematically review the existing approaches to modeling point processes in terms of three families:
simple, Marked , and spatio-temporal point processes, covering evaluation metrics, datasets and application areas.
\item We provide a deep discussion on the main concerns in each family of the approaches. 
\item We 
explore the potential application areas.
\end{itemize}

The paper is organized as follows. Section \ref{sec:background} introduces the background and basic notation, laying the mathematical foundations. In Sections~\ref{sec:tpp}, \ref{sec:mtpp} and  \ref{sec:stpp}  we review Temporal Point Processes, and Marked Temporal Point Process and Spatial Temporal Point Process, respectively. In Section \ref{sec:applications} application areas and datasets are presented. Finally, Section \ref{sec:conclusions} concludes the survey.

\section{Background and Notation} \label{sec:background}

\begin{figure}
    \centering
    \includegraphics[width=.95\textwidth]{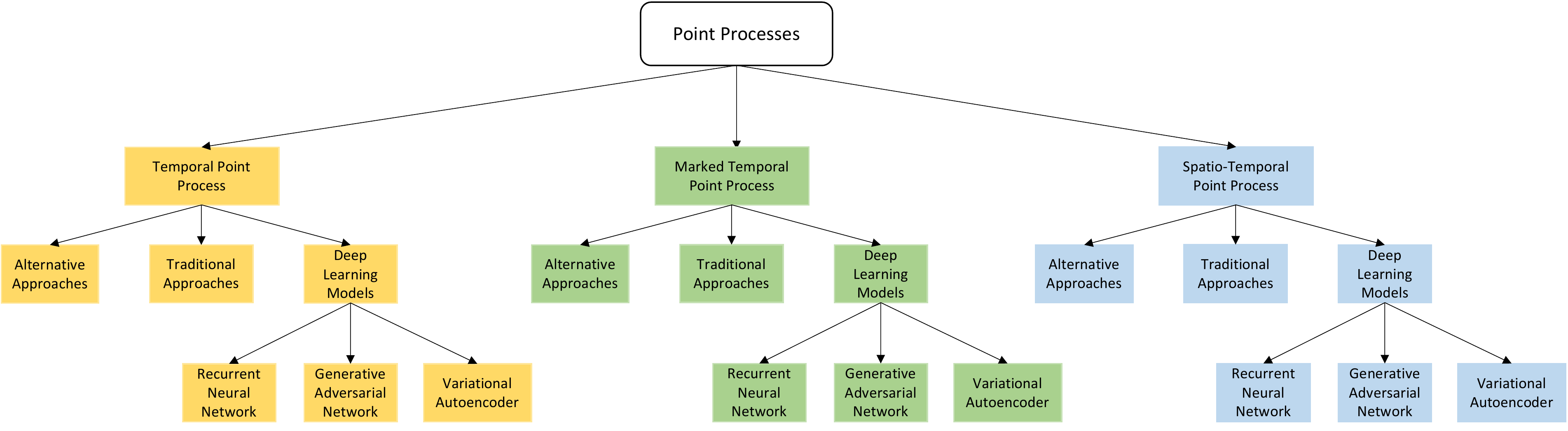}
    \caption{Ontology}
    \label{fig:ontology}
\end{figure}






\subsection{Point Processes}

\textit{Point Processes} (\textit{PPs}) are a mathematical tool for modeling stochastic processes, i.e, sequences of events, in continuous time. Mathematically, each sequence is represented by a set of time-ordered events $ E = \{e_1, e_2, ..., e_N \}$ where $N$ refers to the total sequence length. Based on the information stored in the sequence, the point processes, as shown in Fig. \ref{fig:ontology} can be categorized in: (\textit{i}) \textit{Temporal Point Processes} (\textit{TPPs}) in which each event $e_i$ is characterized by one attribute, i.e, the event timestamp $t_i$, $e_i = t_i$; (\textit{ii}) \textit{Marked Temporal Point Processes} (\textit{MTPPs}), i.e., each event is characterized by two attributes, the event timestamp and the event type, $e_i = (t_i, k_i)$; (\textit{iii}) \textit{Spatio-Temporal Point Processes} (\textit{STPPs}) in which each event stores the information about the time and the location where the event occurs, $e_i = (t_i, s_i) $.

Given a set of event sequences, the aims is to understand the structural and temporal dynamics characterizing them in order to predict future events, i.e., predict when an event will happen in the future and/or its nature in terms of event type and location. The sequence of events has an evolutionary characterization, i.e., what happens in a specific time $t$ depends on what happened in the past, but not on what is going to happen in the future. Our focus is on probabilistic tools to model this evolution. Specifically, event evolution can be modeled by considering the probability of the next event(s), given its history - $P (e_t|H_t)$ where $H_t$ denotes the history composed by all the events happened before the time $t$, i.e., $H_t = \{ e_i \in E| t_i < t\}$. We shall assume that $n=|H_t|$ and   $t_n < t$ represents the time associated with the last event in $H_t$, in chronological order. 

Since in MTPPs, each event is characterized by two attributes, the event timestamp and a categorical attribute representing the type of the event, the probability distribution to learn can be decomposed in two ways: (i) By assuming independent decomposition in which, given the history, we can model the probability of the event type and the probability of time independently; (ii) or, by contrast, assuming dependence and confounding, in which the probabilities of an event type and its time of occurrence are correlated.

Given the history of past events $H_t$, the probability of a Point Process is modeled through a conditional intensity function, which measures the number of events of a given type in an instant of time (and/or space) that can be expected in a time interval $t$ (and/or in a location $s$). 
Based on the nature of the point process, we have different formulations of the conditional intensity function (see the sections below).

\subsubsection{Temporal Point Processes}
A temporal Point Process is modeled by a conditional intensity function that can be expressed in terms of counting process. The latter, denoted by $N(t)$,  reports the number of events occurring at time $t$. Then, given an infinitesimal interval $[t, t+\Delta t$), the conditional intensity function $\lambda(t|H_t) \geq 0$ is the occurrence rate for the future event conditioned on history $H_t$ up to but not including time $t$:

\begin{equation}
    \lambda(t|H_t) = \lim_{\Delta t \rightarrow 0} \frac{\mathbb{E}[N([t, t+\Delta t)) | H_t]}{\Delta t } = \frac{\mathbb{E} [dN(t)|H_t]}{dt}
\end{equation}

Therefore, the conditional intensity function specifies the average number of events in an instant of time, conditioned on the past \cite{paper23}. Equivalently, given the history $H_t$, a Temporal Point Process can be characterized considering the following three functions:
\begin{enumerate}
    \item Conditional density function $f(t|H_t)$: Given the history $H_t$, it represents the probability that the next event occurring in $[t, t+\Delta t)$, for an infinitesimal interval $\Delta t$;
    \item Cumulative Distribution Function $F(t|H_t) = \int_{t_n}^{t} f(\tau|H_{\tau})d\tau$: the probability that the next event will happen before the time $t$, given the history $H_t$;
    \item Complementary Cumulative Distribution Function, also called Survival Distribution, $S(t|H_t) = 1 - F(t|H_t)$: the probability that no event has ever happened up to time $t$.
\end{enumerate}

By considering these functions, the conditional intensity functions is defined as follows:
\begin{equation}
    \lambda (t|H_t) = \frac{f(t|H_t)}{1-F(t|H_t)} = \frac{f(t|H_t)}{S(t | H_t)}
\label{conditional_intensity_function2}
\end{equation}
Starting from one of these functions, it is possible to derive the others. Specifically, the Survival Distribution is defined as $S(t|H_t) = 1 - F(t|H_t) = 1 - \int_{t_n}^{t} f(\tau|H_{\tau})d\tau$. Therefore, we have $dS(t|H_t) = -f(t|H_t)dt$. We can rewrite equation \ref{conditional_intensity_function2} as follows:
\begin{align}
    \begin{split}
    \lambda(t|H_t ) &= \frac{f(t|H_t)}{S(t|H_t)} \\
    &= -\frac{1}{S(t|H_t)} \cdot \frac{dS(t|H_t)}{dt} \\
    &= - \frac{d}{dt} \log S(t|H_t)
    \end{split}
\end{align}
Applying the integration on both the side, we obtain:
\begin{align}
    \begin{split}
    \int_{t_n}^{t} \lambda(\tau|H_{\tau} )d\tau &= - \int_{t_n}^{t} \frac{d}{d\tau} \log S(\tau|H_{\tau}) \\
    &= - \log S(t|H_t)
    \end{split}
    \label{conditional_integration}
\end{align}
From equation \ref{conditional_integration}, we obtain:
\begin{equation}
    S(t|H_t) = \exp\bigg(- \int_{t_n}^{t} \lambda(\tau|H_{\tau}) d\tau\bigg)
\end{equation}
and 
\begin{equation}
    f(t|H_t) = \lambda(t|H_t) \cdot \exp \bigg(- \int_{t_n}^{t} \lambda(\tau|H_{\tau}) d\tau\bigg)
\end{equation}

\begin{figure}
    \centering
    \includegraphics[width=.55\textwidth]{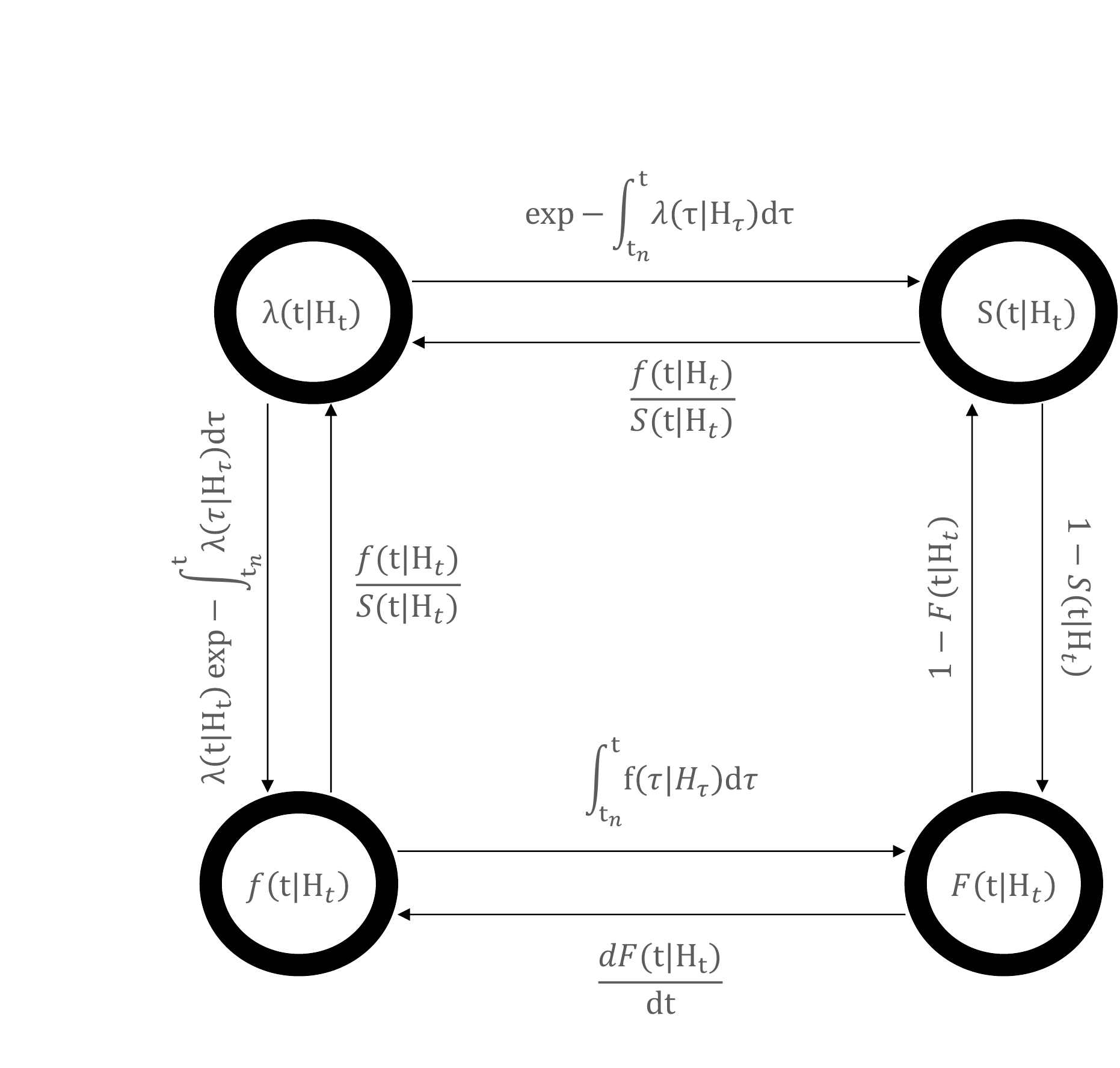}
    \caption{Relation between the conditional intensity function $\lambda(t|H_t)$, the conditional density function $f(t|H_t)$, the cumulative distribution function  $F(t | H_t)$ and the survival distribution $S(t | H_t)$.}
    \label{fig:relation}
\end{figure}

Figure \ref{fig:relation} shows the relation between these functions.

The standard approach to learn a temporal point process consists in parameterizing its intensity function  $\lambda_{\theta}(t | H_t)$ by means of a parameter set $\theta$ and learning the most appropriate instantiations for such parameters. Given a temporal point process $\{ t_1, t_2, ..., t_N \}$ in an interval $[0, T)$ modeled by its conditional intensity function $\lambda_{\theta} (t | H_t)$, the likelihood function can be rewritten as:
\begin{equation}
    \mathcal{L}_{\theta}(H_T) = \prod_{i=1}^N \lambda_\theta (t_i | H_{t_i}) \cdot \exp \bigg( \int_0^T \lambda_\theta (\tau | H_\tau) d\tau \bigg)
\end{equation}
and the log-likelihood is defined as follow:
\begin{align}
\begin{split}
    \ell(\theta) &= \log \mathcal{L}_{\theta}(H_T) \\
    &= \log \bigg(\prod_{i=1}^N \lambda_\theta (t_i | H_{t_i}) \cdot \exp \bigg(- \int_0^T \lambda_\theta (\tau | H_\tau) d\tau \bigg) \bigg) \\
    &= \sum_{i=1}^N \log \lambda_\theta (t_i | H_{t_i}) - \int_0^T \lambda_\theta (\tau | H_\tau) d\tau
\end{split}
\end{align}
The underlying TPP can be hence learned by maximum likelihood estimation on $\ell(\theta)$. 


It is worth mentioning some typical functional forms of the conditional intensity function that have been studied in the literature. A \textit{Poisson process} is the simplest point process in which the conditional intensity function is independent of the history. Specifically, there are two type of Poisson process: (\textit{i}) \textit{Homogeneous Poisson process} where the intensity function is given by a constant parameter, i.e., $\lambda(t) = \mu \geq 0$ and (\textit{ii}) \textit{Inhomogeneous Poisson process} defined by a time-varying function, i.e, $\lambda(t) = g(t) \geq 0$. 

The \textit{Hawkes processes}~\cite{10.2307/2334319,Laub2015HawkesP} represent a specific extension of the Poisson process, where the history exerts a \textit{self-exciting} influence:
\begin{equation}
    \lambda(t | H_t ) = \mu(t) + \alpha \sum_{t_i < t} g(t-t_i)
\end{equation}
Here, the first term represent a background intensity, while the second is the excitation function. The core idea is that, when multiple events happen, the intensity function grows by $\alpha > 0$ (according to the function $g$) and then decreases back towards $\mu$ as soon as the events get far in time.  \textcite{10.2307/2334319} originally used an exponential decay $g(t) = e^{-\beta \cdot t}$ for the excitation function, with the meaning that each arrival increases the intensity by $\alpha$ and decreases its influence exponentially with rate equal to $\beta>0$.

\subsubsection{Marked and Spatio-Temporal Point Processes}
In Marked Temporal Point Processes, the conditional intensity function is the probability of observing an event of type k in an infinitesimal interval $[t, t+\Delta t)$:
\begin{equation}
    \lambda(t, k | H_t ) = \lim_{\Delta t \rightarrow 0} \frac{\mathbb{E} [N([t, t+ \Delta t) \times k) | H_t]}{\Delta t}
\end{equation}
A special case is given by Spatio-Temporal Point Processes, where the focus is on events observed in an infinitesimal around (t, s):
\begin{equation}
    \lambda(t, s | H_t ) = \lim_{\Delta t, \Delta s \rightarrow 0} \frac{\mathbb{E}[N([t, t+\Delta t) \times B(s, \Delta s)]) |H_t] }{\Delta t \times |B(s, \Delta s)|}
\end{equation}
Here, $B(s, \Delta s)$ denotes an euclidean ball centered at $s$ with radius $\Delta s$.

The generalizations to event types or spatio-temporal events applies accordingly, also to the density, distribution and survival functions $f(t|H_t)$, $F(t|H_t)$, $S(t|H_t)$ defined in the previous section.

\section{Temporal Point Processes} \label{sec:tpp}

\begin{table}[]
    \centering
    \resizebox{\linewidth}{!}{
    \begin{tabular}{l|c c c c c c}
         Paper & Data Type & 
         Architecture & 
         Year  \\
         \hline
         
         \makecell[l]{Learning Temporal Point Processes \\ via Reinforcement Learning} \cite{paper25} & Sequence of events & 
         \makecell{Reinforcement Learning Framework \\ where the policy is modeled by \\ a Stochastic RNN} & 
         2018\\ 
         \hline
         
         \makecell[l]{Wasserstein Learning of Deep Generative \\ Point Process Models} \cite{paper26} & Sequence of event times & 
         \makecell{Generative Adversarial Networks, \\ Recurrent Networks} & 
         2017 \\
         \hline
         
         \makecell[l]{Learning Mixture of Neural Temporal Point Processes \\ for Multi-dimensional Event Sequence Clustering} \cite{paper27} & Sequence of events 
         & Deep Neural Network & 
         2022\\
         \hline
         
         \makecell[l]{Deep Recurrent Survival Analysis} \cite{paper29} & \makecell{Survival events both \\ censored and uncensored (x,t,e)} 
         & Recurrent Neural Network & 
         2019\\
         \hline
         
         \makecell[l]{Time2Vec: Learning a Vector Representation of Time} \cite{paper37} &  
         & & 
         2019 \\ 
         \hline
         
         \makecell[l]{Learning Conditional Generative Models for \\ Temporal Point Processes} \cite{paper50} & Sequence of events 
         & \makecell{Wasserstein generative adversarial network \\ with sequence-to-sequence model} & 
         2018 \\ 
         \hline
         
         \makecell[l]{Transformer-Based Deep Survival Analysis} \cite{paper60} &  Sequence of events 
         & Transformer & 
         2021 \\ \hline
         
        \makecell[l]{Calibration and Uncertainty in \\ Neural Time-to-Event Modeling} \cite{paper62} & Sequence of events 
        & Multilayer Perceptron, GAN & 
        2020 \\ \hline
          
         \makecell[l]{Know-Evolve: Deep Temporal Reasoning \\ for Dynamic Knowledge Graphs} \cite{paper67} & Sequence of events 
         & Recurrent neural network & 
         2017 \\ \hline

         \makecell[l]{Adversarial Time-to-Event Modeling} \cite{paper69} & Sequence of events 
         & Generative adversarial network & 
         2018 \\ \hline
         
         \makecell[l]{Point Process Flows} \cite{paper76} & Sequence of events 
         & normalizing flows & 
         2019 \\ \hline
         
         \makecell[l]{Geometric Hawkes Processes with \\ Graph Convolutional Recurrent Neural Networks} \cite{paper80} & Sequence of events
         & \makecell{Graph convolutional network, \\ recurrent neural network} &
         2019 \\ \hline
         
         \makecell[l]{Temporally-Consistent Survival Analysis} \cite{paper82} & Sequences of states  
         & \makecell{temporally-consistent survival regression \\ in which the survival distribution is modeled \\ in a recursive way under the \\ Markov assumption} &
         2022 \\ \hline

         \makecell[l]{Variational Policy for Guiding Point Processes} \cite{paper83} 
         &  Sequence of events 
         & Stochastic Differential Equations (SDEs) 
         & 
         2017 \\
         \hline

        \makecell[l]{Cheshire: An Online Algorithm for \\ Activity Maximization in Social Networks} \cite{paper84} & 
        Graphs
        & Stochastic Differential Equations (SDEs) & 
        2017 \\
        \hline

        \makecell[l]{Learning and Forecasting Opinion Dynamics \\ in Social Networks} \cite{paper85} & 
        Graph 
        & 
        Stochastic Differential Equations (SDEs) & 
        2015 \\
        \hline

        \makecell[l]{A Dirichlet Mixture Model of Hawkes Processes for \\ Event Sequence Clustering} \cite{paper86} & 
        Sequence of events 
        & 
        Expectation Maximization (EM) & 
        2017 \\
        \hline

         \makecell[l]{Counterfactual Phenotyping with \\ Censored Time-to-Events} \cite{paper91} & Time series & 
         MLP &
         2022 \\ \hline
         
         \makecell[l]{Continual Learning for \\ Time-to-Event Modeling} \cite{paper92} 
         & Sequence of events 
         & \makecell{Neural Hawkes Process with \\ a meta-network} &
         2022 \\ \hline
           
    \end{tabular}}
    \caption{Temporal Point Processes}
    \label{tab:tpp_table}
\end{table}			

\subsection{Deep Learning Methods}

\subsubsection{Recurrent Neural Networks}

In \cite{paper25}, the authors propose a Reinforcement Learning (RL) framework to learn point process models. It processes sequences of events occurred at instants of times (belonging to a continuous timeline) $H_t=\{t_i\ |\ i\in[1..N], t_i<t\}$, called \emph {trajectories}.
The observed trajectories are regarded as sequences of actions taken by an expert which is characterized by an unknown policy $\pi_E$. The goal is to learn a policy $\pi(d\ |\ H_t)$ that mimics the distribution of the observed expert action sequences. Here, $d$ is an inter-event time representing the prediction that the next event will occur at time $t_N + d $. The policy is modeled by a Recurrent Neural Network (RNN) with stochastic neurons, which can sequentially sample discrete events. Different from traditional RNNs, the outputs of a stochastic RNN are not deterministic but are sampled from a specific distribution whose parameters are computed by the model. Moreover, each sampled output is fed back to the RNN and will be used to compute its next hidden state. The use of a stochastic RNN is a way to model an exploration/exploitation strategy needed in a RL architecture. It is trained by a reinforcement learning approach, directly minimizing the discrepancy between the generated sequences and the observed sequences. In this framework, the discrepancy is explicitly evaluated in terms of the reward function. Maximizing the reward will iteratively encourage the policy to sample events as close as possible to the observation.
\\

In \cite{paper29} a DL-based model for survival analysis is proposed. The model combines recurrent networks and survival analysis to respectively model the conditional probability and to handle the censorship. The model is dubbed DRSA (Deep Recurrent Survival Analysis). Authors argue that this framework not only proves to be competitive and accurate, but also provide higher flexibility since it does not require to fit any underlying parametric distribution. The model design treats the time as a discrete variable using RNNs - specifically a Long-Short Term Memory (LSTM). Each event of the sequence is given as input to an LSTM, and the extracted features are used to estimate the conditional risk probability at the event time. In the end, the conditional risk probability at the end of the sequence is based on the features of all the previous events. Basic block of this architecture is a RNN - specifically, an LSTM - which is used to extract relevant information to estimate the conditional risk probability. Each state of the RNN - in this case, the classical LSTM - provides the probability at a specific time range. This design choice treats the time as a discrete state, and the prediction at each time range is given by the features derived by all the previous states. For this reason, the authors state that DRSA is able to take into account the sequential patterns in the feature space.
\\

Since time is an important feature for synchronous and asynchronous event prediction tasks, in \cite{paper37}, the authors propose to represent the time feature by means of a vector - dubbing their approach time2vec. The presented approach allows for more flexibility since it can accommodate three main objectives: (\textit{i}) Periodicity, (\textit{ii}) invariance to scaling and (\textit{iii}) simplicity. Since it is model-agnostic, it can be imported in any existing and future architecture. In the experimental analysis, authors integrates \textit{time2vec} in a LSTM architecture and they show how this improves the performance of the model on both synthetic and real data. They also show how it effectively realizes the objectives of periodicity, and invariance.
\\

The goal of \cite{paper67} is to predict the occurrence of a fact, i.e, an edge in a temporal dynamic graph, and when the fact may occur. Specifically, the graph is defined as a quadruplet ($e^s$, $r$, $e^o$, $t$) where $e^s$ and $e^o$ represent the subject and object entities, i.e., the source and the destination. The quadruplet represents the creation of relationship edge $r$ between the subject and the object entities at time $t$. The occurrence of these events is modeled by using the conditional intensity function: $ \lambda_{r}^{e^s, e^o}(t|\hat{t})$ where $t$ is the time of the current event and $\hat{t} = max(t^{e^s}, t^{e^o}) $ is the most recent time when either subject or object was involved in an event before $t$. This function is modulated by the score between the involved entities in that specific relationship computed based on the learned entity embeddings. Each embedding is computed by using a recurrent neural network. The model is trained by minimizing the joint negative log likelihood of the intensity function. The authors report experimental results on two tasks: Link prediction and time prediction. For the link prediction task, the aim is to predict the object entities. Mean absolute rank, its standard deviation and HITS@10 are used as metrics. Regarding the time prediction, the Mean Absolute Error (MAE) is used as a metric. The results show that the proposed model outperforms the state-of-the-art approaches.
\\

In \cite{paper80}, the authors propose a framework to model temporal events generating the parameters of the Hawkes processes using Graph Convolutional Networks (GCNs) and RNNs. They propose two types of their model: (\textit{i}) Single-graph that is useful to model sequences with one type of graph (user/item graph) and (\textit{ii}) multi-graph that is useful when there are multiple graphs (user graph and item graph). Specifically, the learned embeddings from the combination of GCN and RNN are used to generate the parameters of the Hawkes processes. The log-likelihood with regularizers is used as a loss function. Several baseline and three real-world datasets are used to perform the experiments. As metrics, the authors use the mean average rank to show the item relevance and the mean absolute error for the time prediction task. The experiments show that the proposed model outperforms state-of-the-art approaches and that the multi-graph version is better w.r.t. the single one since it is able to extract more information.
\\

\subsubsection{Generative Models (VAE and GANs)}
In \cite{paper26} the authors tackle the problem of predicting when the next event will happen. Starting from the limitations of current approaches based on point processes, they depict an intensity-free approach to solve two main issues: First, unrealistic assumptions of parametrical models, and second, the learning process through maximum likelihood estimation that fails when the distribution is inherently multi-modal. For this reason, they design a framework that leverages the min-max strategy of Generative Adversarial Networks (GANs) to optimize a recurrent neural network to generate sequences with high fidelity to the ground truth. Hence, they adopt Wasserstein distance to measure how much the generative module can replicate distributions similar to the real ones. Finally, theoretical analysis and experiments on synthetic and real data show the superiority of their approach w.r.t. standard point-process methods in the literature.
\\

Given an observed time-dependent event sequence, the goal is to predict the next times of the sequence. The authors in \cite{paper50} use a Wasserstein generative network in which the generator follows the sequence-to-sequence architecture. In particular, the sequence-to-sequence model exploits recurrent neural networks. The observed time-dependent event sequence is fed into the generator that outputs the next time. Specifically, the aim of the generator is to estimate the probability of the next time, given the history and the recurrent neural network iteratively generates the outputs by selecting the value that maximizes that probability. The discriminator, exploiting a Convolutional Neural Network (CNN), aims to distinguish real and generated sequences. The effectiveness of the proposed framework is shown by using both real and synthetic datasets and the model is compared with several baseline and state-of-the-art methods. As a metric, the authors define the Pr.Dev. metric that compares the error over the predicted sequence with the real sequence. The experiments show that the proposed model outperforms the competitors.
\\

The authors in \cite{paper62} address the problem of \textit{time-to-event}, i.e., the elapsed time until an event of interest occurs, prediction in the domain of survival analysis. Previously, time-to-event distributions were frequently represented using a small set of parametric forms. In practice, these assumptions are frequently violated, resulting in models that are unable to capture unobserved (nuisance) variation. Furthermore, traditional survival analysis applications usually concentrate on a single time-to-event outcome. In practice, events can be caused by a variety of factors, known as competing risks, some of which may be relevant. For these reasons, the authors present approaches that implicitly define time-to-event distributions conditioned on covariates via a neural network specification, from which they can synthesize temporally accurate, concentrated, and calibrated time-to-event distributions. In particular, three neural network-based models are proposed: DRAFT, an Accelerated Failure Time (AFT) inspired model that serves as a baseline; DATE \cite{paper69}, a non-parametric model that leverages adversarial learning; and SFM, an extension of DATE that replaces the time-to-event distribution matching discriminator with a calibration objective. The qualitative and quantitative experiments compare the proposed approaches against state-of-the-art competitors on five real-world datasets. In terms of estimating concentrated and calibrated time-to-event distributions, SFM performs better than other approaches while maintaining a competitive C-index (metric used for the evaluation).
\\

In \cite{paper69}, the authors provide a non-parametric approach based on generative adversarial network to model the time-to-event. Their model aims at estimating the probability of the time-to-event given the observed covariates $X$. Since the time-to-event could be not observed, the authors consider the non-censored and censored time-to-event. The model (called DATE) consists in a generator (G) and a discriminator (D) network, both specified as a deep neural network. The generator takes as input the covariates, a random noise and the indicator and provides as output the generated time-to-event. The discriminator aims to distinguish if the time is the real time or the generated ones. For non-censored time, the loss function is defined as:
\begin{equation}
    l_1 = E_{(t, x) \sim p_{nc}} [D(x, t)] + E_{x \sim p_{nc}, \epsilon \sim p_\epsilon} [1-D(x, G(x, \epsilon, l=1)]
\end{equation}

For the censored data, an addition loss function is added: 
\begin{equation}
    l_2 = E_{(t, x) \sim p_c, \epsilon \sim p_\epsilon} [max(0, t-G(x, \epsilon; l=0)]
\end{equation}

In addition, in the case in which the proportion of observed events is low, to avoid that the costs of the two losses inderrepresent the desire that time-to-events must be closes as possible to the ground truth, a “distance” loss is added. For the experimental part, the authors also provide a parametric, non-adversarial baseline with a parametric log-normal distribution on the time of event (DRAFT). Experimental results on several real-world datasets show that the proposed model outperforms the baseline. In particular, the authors performed qualitative tests in which they compare the time-to-event distributions and quantitative tests in terms of absolute error relative to the event range ($ \frac{|\hat{t} - t |}{t_{max}}$ for non-censored, $
\frac{\max (0, t-\hat{t})}{t_{max}}$ for censored). A modified version of their model is also provided in which the generator is an adversarial auto-encoder: That version is used to show the robustness of their model to missing data.
\\

\subsubsection{Neural Networks}

In \cite{paper27} a mixture of neural temporal point processes (NTPP-MIX), a general framework that can incorporate various Maximum Likelihood Estimation-based NTPPs for multi-dimensional event sequence clustering is proposed. Specifically, the prior mixing coefficients are modelled by a Dirichlet distribution. Then the coefficients are used to model the prior cluster assignment. For each sequence, its conditional probability on a given assignment is modelled via a series of NTPPs.
The variational framework, NTPP-MIX, has a natural tendency to set some mixture weights close to zero and automatically choose a suitable number of components. When \textit{K} clusters are unknown, we initialize \textit{K} as a large number and remove redundant clusters during training. In terms of the evaluation protocol, the authors used several synthetic and real-world datasets to assess the performance of the proposed method. Several baselines are used for comparative analysis.
\\

In \cite{paper60}, the authors address the problem of estimating patient-specific survival distributions. The paper's novelty is applying transformers, commonly used in Natural Language Processing (NLP) tasks, to survival analysis. For this purpose, each patient is treated as a sentence, and each word represents the interaction between the patient and time. Ordinal regression is also used to optimize survival probabilities over time and to penalize randomized discordant pairs. Furthermore, unlike previous approaches, the authors propose evaluating the results using both the C-index and the mean absolute error. The C-index is a ranking metric, not a duration evaluation metric, so inaccurate models could still receive a high score. The method is tested on two real-world medical datasets and compared to four baseline models. The proposed approach is significantly better in terms of MAE and comparable in terms of the C-index.
\\

The authors of \cite{paper91} propose a principled approach, Cox Mixtures with Heterogeneous Effects, to discover subgroups or cohorts of individuals that demonstrate heterogeneous effects to an intervention in the presence of censored outcomes. The proposed method is not sensitive to strong assumptions of proportional hazards and can be applied in situations where the effect of a treatment is not uniform across the population. The features (confounders) are passed through an encoder to obtain deep non-linear representations. These representations then describe the latent phenogroups that determine the base survival rate and the treatment effect, respectively. Finally, the individual-level hazard (survival) curve under intervention A is determined. To show the effectiveness of the proposed model, the authors use several datasets and compare their model with two baseline methods.
\\

In \cite{paper92}, the authors tackle the problem of catastrophic forgetting in time-to-event modeling developing "HyperHawkes", a model that exhibits a meta-network that computes the parameters used by a fully neural Hawkes point process, and a regularization term on the learned parameters that penalize any change w.r.t. previous model snapshots to retain prior knowledge. Experimentation on two real-world datasets demonstrates the effectiveness of the proposed framework to provide both higher accuracy and less forgetting w.r.t. a fully-neural Hawkes process.
\\

\subsection{Traditional Statistical Methods}

In \cite{paper82}, the authors propose an algorithm that estimates a survival model in the dynamic setting.  Specifically, in the dynamic setting, the data consist of sequences of states (measurements over time) and the authors assume that those sequences follow a Markov chain in which the event of interest is modeled as a terminal state. The goal is to estimate the survival distribution, i.e., the distribution of hitting time for that terminal state (the time-to-event in survival analysis annotation), from any other state. Under the Markov assumption, the survival probability at a given state should be similar (on average) to the survival probability at the next state ($S(k|x) E_{x' \sim p(\cdot|x)} [S(k-1 | x')] $ where $p(\cdot|x)$ is the transition probabilities) and in this way the survival probability can be computed in a recursive way. The authors begin by assuming that the Markov chain dynamics are known and present a dynamic programming approach to compute the exact survival distribution from every state. Then, they extend the approach to fit a parametric approximation to the survival distribution. The authors argued that in real-world scenarios, the transition probabilities are not known, therefore they propose a temporally-consistent survival regression (TCSR) algorithm that estimates a survival model from samples. In the experimental evaluation, the authors compare the proposed method with the model only using initial-state information and the “landmark” approach in which the initial-state approach is extended by including every intermediate state and the remaining time-to-event. The authors use both synthetic and real-world datasets showing that the proposed method outperforms the baseline in terms of concordance index and Root Mean Squared Error (RMSE).
\\

The topic of \cite{paper83} is the \textit{stochastic system steering}. The authors propose a generic framework for controlling the stochastic intensity function of a general point process so that a non-linear stochastic differential equation driven by the point process is steered toward a target state. A network moderator, for example, may want to influence user behavior to prevent the spread of false rumors or encourage the posting of educational topics. In order to achieve the desired state, this work aims to identify the best control policy. The novelty of the approach is threefold. It provides a generic method for controlling non-linear stochastic differential equations driven by stochastic point processes. In contrast to previous works, no approximations of the system or the objective function are required. It is not necessary to design a meaningful control cost function to optimize because this function emerges naturally from the structure of the stochastic dynamics. It is fast, scalable, and parallelizable. The study's key finding is that the stochastic optimal control problem can be viewed from the angles of optimal measure and variational inference. The method is validated using synthetic and real-world datasets representing two tasks: Guiding opinion diffusion and broadcasting behavior. Performance is compared with suitable stochastic optimization approaches that are popular in reinforcement learning and heuristics. Experiments show that the proposed algorithm is more accurate and efficient than other stochastic control methods in steering user activities.
\\

The authors in \cite{paper84} propose Cheshire, an efficient online algorithm that incentivizes actions to maximize users' overall network activity. There are two major differences between this and previous works. First, the control signal used to sample the times of incentivized actions is viewed as a multidimensional conditional intensity rather than a time-varying real vector. Second, user intensities are regarded as stochastic Markov processes rather than deterministic. The authors use a temporal point process framework to represent user behavior. This framework uses conditional intensity functions to characterize the continuous time interval between actions and multidimensional Hawkes processes to model endogenous and exogenous actions. Then, using Stochastic Differential Equations (SDEs) with jumps, they derive an alternate representation of multidimensional Hawkes processes. Using this alternate representation, they cast the activity maximization problem as a novel optimal control problem. Cheshire is tested using both synthetic and real Twitter data. The approach outperforms the state-of-the-art in terms of consistently maximizing the number of actions.
\\

The authors in \cite{paper85} propose SLANT, a data-driven model for opinion dynamics capable of forecasting user opinions. SLANT is a probabilistic framework for representing user opinions over time by employing continuous-time stochastic processes driven by a set of marked jump SDEs. This enables efficient model simulation and parameter estimation from fine-grained historical event data.
One of the primary shortcomings of competing approaches is that they do not distinguish between latent and expressed opinions (or sentiment). In this work, each user's latent opinion is modulated over time by the sentiment messages expressed asynchronously by its neighbors. The approach is tested on both synthetic and real data gathered from Twitter. The authors demonstrate that SLANT provides an excellent fit to the data and that the predictive formulas achieve more accurate opinion forecasting than a number of alternatives. The approach can result in a variety of opinion dynamics, which may or may not lead to a steady state of consensus or polarization. Furthermore, the model estimation, simulation algorithms, and predictive formulas proposed scale to networks with millions of users and events.
\\

The authors in \cite{paper86} propose DMHP (Dirichlet Mixture Model of Hawkes Processes), a clustering method for point process sequences. This method models the event sequences from different clusters using different Hawkes processes. The Hawkes process parameters' priors, in particular, are designed based on physically meaningful constraints, whereas the prior of the clusters is generated using a Dirichlet distribution. The proposed algorithm performs a variational Bayesian inference to learn the DMHP model in a nested Expectation-Maximization (EM) framework. Furthermore, using open-loop control theory, the authors incorporate a novel inner iteration allocation strategy into the algorithm, which improves the algorithm's convergence. This is the first systematic study of the identifiability problem in event sequence clustering. The authors formally demonstrate DMHP's local identifiability and conduct experiments on synthetic and real-world datasets. They show that the proposed clustering method can robustly learn structural triggering patterns hidden in asynchronous event sequences and achieve superior performance on cluster purity and consistency compared to existing methods.
\\

\subsection{Alternative Approaches}
The authors in \cite{paper76} describe an intensity-free framework to model the point process distribution via normalizing flows. This technique has the main advantage of not requiring any explicit parametric form, and it can model complex time distributions and perform predictions. Furthermore, the proposed model - dubbed Point Process Flow - is optimized by maximizing the likelihood using change of variable formula and relaxing the strict tractable likelihood of previous works.

\section{Marked Temporal Point Processes} \label{sec:mtpp}
        
\begin{table}[]
    \centering
    \resizebox{0.86\textwidth}{!}{
    \begin{tabular}{l|c c c c c c}
         Paper & Data Type & 
         Architecture & 
         Year  \\
         \hline
         \makecell[l]{Uncertainty-Aware \\ Anticipation of Activities} \cite{paper1} & 
         Sequence of video frames & 
         recurrent neural network & 
         2019 \\
         \hline
         
         \makecell[l]{A Variational Auto-Encoder Model \\ for Stochastic Point Processes} \cite{paper2} &
         Sequence of actions &
         Variational Autoencoder &
         2019 \\
         \hline
         
         \makecell[l]{Recurrent Marked Temporal Point Processes: \\ Embedding Event History to Vector} \cite{paper3} &
         Sequences of events (Marker + Time) & 
         Recurrent Neural Network & 
         2016\\
         \hline
         
         \makecell[l]{When will you do what? - \\ Anticipating Temporal Occurrences of Activity} \cite{paper4} & 
         sequence of video frames &
         Convolution Neural Network and Recurrent Neural Network &
         2018 \\
         \hline
         
         \makecell[l]{Predictive Business Process Monitoring \\ via Generative Adversarial Nets: \\ The Case of Next Event Prediction} \cite{paper5} & 
         Sequence of events (marker + time) &
         Generative Adversarial Network & 
         2020 \\
         \hline
         
         \makecell[l]{Encoder-Decoder Generative Adversarial Nets \\ for Suffix Generation and Remaining Time Prediction \\ of Business Process Models} \cite{paper6} & 
         
         \makecell{Process logs, i.e. time-stamped event \\ sequences}  & 
         
         \makecell{Generative Adversarial Networks with \\ Recurrent Neural Networks} 
         &
         2020  \\
         \hline
         
         \makecell[l]{An empirical comparison of deep-neural-network architectures \\ for next activity prediction using \\ context-enriched process event logs} \cite{paper7} &
         Process logs, i.e. time-stamped event sequences & 
         MLP, RNN, CNN &
         2020 \\
         \hline
         
         \makecell[l]{What Averages Do Not Tell - \\ Predicting Real Life Processes with \\ Sequential Deep Learning}  \cite{paper8} & 
         Sequence of events & 
         Deep Learning & 
         2022  \\
         \hline
         
         \makecell[l]{A Deep Adversarial Model for \\ Suffix and Remaining Time Prediction \\ of Event Sequences} \cite{paper9} & 
         Process logs, i.e. time-stamped event sequences &
         \makecell{Generative Adversarial Networks with \\ Recurrent Neural Networks} 
         & 
         2021 \\
         \hline
         
         \makecell[l]{What Happens Next? Event Prediction \\ Using a Compositional Neural Network Model} \cite{paper10} & 
         Text sequences
         & 
         word2vec and DNN &
         2016 \\
         \hline
         
         \makecell[l]{Attentive Neural Point Processes \\ for Event Forecasting} \cite{paper11} & 
         Sequence of events 
         &
         neural network with attention & 
         2021 \\
         \hline
         
         \makecell[l]{Deviation-based Marked Temporal \\ Point Process for Marker Prediction} \cite{paper13} & 
         Sequence of events & 
         Recurrent neural network & 
         2021 \\
         \hline
         
         \makecell[l]{SEISMIC: A Self-Exciting Point Process Model \\ for Predicting Tweet Popularity} \cite{paper14} & 
         Sequence of events with side information (node degrees) &
         Statistical model & 
         2015 \\
         \hline
         
         \makecell[l]{Uncertainty on Asynchronous \\ Time Event Prediction} \cite{paper15} & 
         Sequence of events & 
         RNN &
         2019 \\
         \hline
         
         \makecell[l]{A Variational Point Process Model \\ for Social Event Sequences} \cite{paper16} & 
         Sequence of events &
         LSTM + VAE &
         2020\\
         \hline
         
         \makecell[l]{What is More Likely to Happen Next? \\ Video-and-Language Future Event Prediction} \cite{paper17} & 
         Sequence of events &
         Autoencoder & 
         2020\\
         \hline
         
         \makecell[l]{Variational Neural Temporal Point Process} \cite{paper18} & 
         Sequence of events & 
         Variational Autoencoder with Transformer & 
         2022 \\
         \hline
         
         \makecell[l]{Future Event Prediction: If and When} \cite{paper19} & video data 
         & \makecell{3D-ResNet for representing video excerpts, \\ generic DNN to predict output} & 
         2019 \\
         \hline
         
         \makecell[l]{Wasserstein generative adversarial networks \\ for modeling marked events} \cite{paper20} & 
         Sequence of events & 
         Wasserstein Generative Adversarial Network & 
         2022 \\
         \hline
         
         \makecell[l]{Deep Reinforcement Learning of \\ Marked Temporal Point Processes} \cite{paper24} & 
         Sequence of events & 
         Deep Q-Learning where the Deep Q-Networks are LSTMs & 
         2018 \\ \hline

         \makecell[l]{Deep Structural Point Process for \\ Learning Temporal Interaction Networks}  \cite{paper30} & 
         Temporal interaction network  & 
         Deep Neural Networks, Recurrent networks, Attention & 
         2021 \\ \hline
         
         \makecell[l]{Tracing temporal communities and event prediction \\ in dynamic social networks} \cite{paper31} & 
         Sequence of user actions &
         \makecell{Any (the authors proposed a general methodology that \\ can be used with all the classifiers) but they used MEKA.} & 
         2019 \\ \hline
         
         \makecell[l]{A Model-Free Approach to Infer \\ the Diffusion Network from Event Cascade}  \cite{paper32} &
         Sequence of events & 
         Clustering algorithm & 
         2016 \\ \hline
         
         \makecell[l]{Intensity-Free Learning of \\ Temporal Point Processes} \cite{paper34} & 
         Sequences of event & 
         Recurrent neural network with normalizing flows &
         2020 \\ \hline
         
         \makecell[l]{Self-Attentive Hawkes Process} \cite{paper35} & 
         Sequence of events & 
         self-attention model &
         2020 \\ \hline
         
         \makecell[l]{Transformer Hawkes Process} \cite{paper38} & 
         Sequences of events & 
         Transformer model &
         2020 \\ \hline
         
         \makecell[l]{The Neural Hawkes Process: A Neurally \\ Self-Modulating Multivariate Point Process} \cite{paper39} & 
         Sequences of events & 
         LSTM & 
         2017 \\ \hline
         
         \makecell[l]{Modeling the Intensity Function of \\ Point Process Via Recurrent Neural Networks} \cite{paper40} & 
         Sequences of events, time series & 
         RNN & 
         2017 \\ \hline
         
         \makecell[l]{Neural Survival Recommender} \cite{paper41} & 
         Sequence of events & 
         LSTM & 
         2017 \\ \hline
         
         \makecell[l]{Learning Time Series Associated \\ Event Sequences With Recurrent Point Process Networks}  \cite{paper42} & 
         Sequence of events & 
         RNN & 
         2019 \\ \hline
         
         \makecell[l]{DeepDiffuse: Predicting the 'Who' \\ and 'When' in Cascades} \cite{paper43} & 
         Graph & 
         LSTM with attention mechanism & 
         2018 \\ \hline
         
         \makecell[l]{Inf-VAE: A Variational Autoencoder Framework \\ to Integrate Homophily and Influence in \\ Diffusion Prediction} \cite{paper44} &
         \makecell{sequences of pairs (v, t), \\ where v is an infected node and \\ t is the infection time} & 
         VAE with attention mechnism &
         2020 \\ \hline
         
         \makecell[l]{Exploiting Marked Temporal Point Processes \\ for Predicting Activities of Daily Living} \cite{paper45} &
         Sequence of events &
         RNN & 
         2020 \\ \hline
         
         \makecell[l]{Modeling Event Propagation via \\ Graph Biased Temporal Point Process} \cite{paper46} & 
         Sequences of events propagating over a graph. & 
         GNN + RNN &
         2020 \\ \hline
         
         \makecell[l]{Learning Neural Point Processes with Latent Graphs}  \cite{paper47} & 
         Sequence of events & 
         Random Graph and Bilevel Programming & 
         2021 \\ \hline

         \makecell[l]{COEVOLVE: A Joint Point Process Model for \\ Information Diffusion and Network Co-evolution}  \cite{paper53} & 
         Social networks links and messages & 
         Hawkes processes and survival processes & 
         2015 \\ \hline
         
         \makecell[l]{Topological Recurrent Neural Network \\ for Diffusion Prediction} \cite{paper54} & 
         Graph & 
         LSTM &
         2017 \\ \hline
         
         \makecell[l]{Cascade Dynamics Modeling with \\ Attention-based Recurrent Neural Network} \cite{paper55} & 
         Sequence of events & 
         RNN & 
         2017 \\ \hline
         
         \makecell[l]{Time-Dependent Representation for \\ Neural Event Sequence Prediction} \cite{paper56} & 
         Sequence of events &
         RNN & 
         2017 \\ \hline
         
         \makecell[l]{Modeling Sequential Online Interactive Behaviors \\ with Temporal Point Process} \cite{paper57} & 
         Sequence of events & 
         LSTM 
         &
         2018 \\ \hline
         
         \makecell[l]{Marked Temporal Dynamics Modeling Based on \\ Recurrent Neural Network} \cite{paper58} & 
         Sequence of events & 
         RNN & 
         2017 \\ \hline

         \makecell[l]{Temporal Logic Point Processes} \cite{paper63} & 
         Sequence of events 
         & 
         Temporal Logic Programs & 
         2020 \\ \hline
         
         \makecell[l]{Recurrent Point Processes for \\ Dynamic Review Models} \cite{paper64} & Sequences of object reviews & 
         LSTM &
         2020 \\ \hline
         
         \makecell[l]{How Can Our Tweets Go Viral? \\ Point-Process Modelling of Brand Content}  \cite{paper65} & 
         Sequence of events 
         &
         Multivariate, Mutually-exciting Hawkes Process &
         2022 \\ \hline
         
         \makecell[l]{Modeling Marked Temporal Point Process \\ Using Multi-relation Structure RNN}  \cite{paper66} & 
         Sequence of events 
         &
         Recurrent neural network & 
         2020 \\ \hline
         
         \makecell[l]{A gan-based framework for \\ modeling hashtag popularity dynamics \\ using assistive information} \cite{paper68} & 
         Sequences of events with side information &
         ~ & 
         2020 \\ \hline
         
         \makecell[l]{Forecasting Future Action Sequences \\ with Neural Memory Networks} \cite{paper70} & 
         Sequence of events &
         LSTM &
         2019 \\ \hline
         
         \makecell[l]{Semi-supervised Learning for \\ Marked Temporal Point Processes}  \cite{paper71} & 
         Sequence of events 
         & 
         \makecell{semi-supervised model: \\ recurrent neural network (LSTM) + \\ MLP for labeled branch, RNN-based encoder decoder \\ for unlabeled branch} & 
         2021 \\ \hline
         
         \makecell[l]{INITIATOR: Noise-contrastive Estimation for \\ Marked Temporal Point Process} \cite{paper72} 
         & Sequence of events &
         LSTM + NCE &
         2018 \\ \hline
         
         \makecell[l]{Modeling Continuous Time Sequences with \\ Intermittent Observations using \\ Marked Temporal Point Processes}  \cite{paper73} & 
         Sequence of events & 
         RNN &
         2022 \\ \hline
         
         \makecell[l]{ProActive: Self-Attentive Temporal Point Process Flows \\ for Activity Sequences} \cite{paper77} & 
         Sequence of events & 
         RNN &
         2022 \\ \hline
         
         \makecell[l]{Time is of the Essence: A Joint Hierarchical RNN and \\ Point Process Model for Time and Item Predictions}  \cite{paper79} & 
         Sequence of events &
         RNN &
         2019 \\ \hline
         
         \makecell[l]{Modeling the Dynamics of Learning Activity on the Web} \cite{paper87} & Sequence of events &
         Hierarchical Dirichlet Hawkes process & 
         2017 \\ \hline
         
         \makecell[l]{Uncovering Causality from Multivariate \\ Hawkes Integrated Cumulants} \cite{paper88} & 
         Sequence of events & 
         Deep multilayer neural networks &
         2017 \\ \hline

         \makecell[l]{Decoupled Learning for Factorial \\ Marked Temporal Point Processes} \cite{paper89} & 
         Sequence of events 
         & 
         FISTA-based method Logistic Regression &
         2018 \\ \hline

         \makecell[l]{Explainable Hyperbolic Temporal Point Process \\ for User-Item Interaction Sequence Generation} \cite{paper90} & 
         Sequence of preferences & 
         Hyperbolic Temporal Point Process & 
         2022 \\ \hline

         \makecell[l]{Predicting activities of daily living via temporal point processes: \\ Approaches and experimental results} \cite{paper93} & Sequence of events &
         RNN &
         2021 \\ \hline

    \end{tabular}}
    \caption{Marked Temporal Point Processes}
    \label{tab:mtpp_table}
\end{table}

\subsection{Deep Learning Methods}

\subsubsection{Recurrent Neural Networks}

The goal of \cite{paper1} is to predict what happens in the future, predicting all the possible evolution of the sequence taken into consideration: In this way, it is possible to model the uncertainty. The proposed framework aims at modeling the probability distribution of the future activities and uses this to sample several possible sequences of future activities. In addition, the model not only predicts the next activity, but also the length of the next activity: That information is conditioned on the first (next activity). The proposed model consists of two components: The first is the action model that predicts a probability distribution of the future action given the sequence of observed action segments and their lengths, while the second is the length model that predicts a probability distribution of the future action length given the sequence of observed action segments, their lengths, and the future action. For the action model, the authors use a combination of fully connected layer and recurrent neural network (Gated Recurrent Unit (GRU) is instantiated); at the end, they use a softmax layer to get the probability distribution. The probability distribution of the future action length is modeled with a Gaussian distribution: Here, a combination of fully connected layer and recurrent neural network are used to predict the mean and the standard deviation used to model the gaussian distribution. Given the two probabilities distribution, it is possible to generate the future activities. At test time, the authors use two strategies: The first generate multiple sequences of activities by sampling from the predicted distributions (in this way, it is possible manage the uncertainty), in the second strategy a single prediction is taken, by choosing the action label with the highest probability – it is useful for comparison scope since the competitor can predict a single activity. For the evaluation, the authors use two benchmarks, and they report the mean over classes as evaluation metric. Several state-of-the-art approaches are used as competitors: Respect to the state-of-the-art approaches, the proposed model shows a lower accuracy since their model is trained to predict multiple sequences (therefore, they report another metric that is top-1 Mean over Classes (MoC) – considering the top-1 MoC, the proposed model outperforms the state-of-the-art models).
\\

Given a sequence of events in which each event stored the time at which the event happens and the type of the event, the goal of \cite{paper3}  is to predict what kind of event will take place and at what time in the future it occurs. To solve this problem, the authors propose a recurrent neural network to simultaneously model the time and the type. Specifically, their idea is based on the learning of a general representation of the intensity function: It is defined as a nonlinear function of the history, and it is parameterized using a recurrent neural network. The advantage of learning a general nonlinear dependency over the history is that the model can be used to model any family of point processes without prior knowledge of them. Exploiting recurrent neural networks, it is possible to capture the history using the hidden layer of the network: Specifically, the hidden state represents the influence of the history up to the event considered. Given the learned representation, the authors model the marker generation with a multinomial distribution. In addition, they also define the conditional intensity function as a combination of three terms: the first represents the accumulative influence from the marker and the timing information of the past events, the second the influence of the current event and the third is used to give a base intensity level for the occurrence of the next event. In addition, to have a nonlinear function, the authors decide to use an exponential function that also guarantees that the intensity is positive. Given the intensity function, it is possible to estimate the timing for the next event. To evaluate the model, the authors decide to use both synthetic and realistic datasets and the metrics used are the RMSE for the time and the classification error for the marker. The experiments on synthetic datasets want to show that even if the proposed model has no prior knowledge about the true functional form of each process, it is consistent with the result provided by an optimal estimator that knows the true intensity function. In addition, the experiments on both synthetic and realistic datasets show that the proposed model has better performance compared to the baselines.
\\

In \cite{paper4}, the authors address the problem of anticipating activities that will be happening within a time horizon. Specifically, given a sequence of video frames, the goal is to predict which activities happening in the next frames. To address this problem, the authors propose two methods, one based on recurrent neural networks and the other on convolutional neural networks. The first model exploits a recurrent neural network that given a sequence of events, predicts which is the next event as well as the remaining length of the last observed frame and a length for the next frame. This prediction is concatenated with the observed segments to form the new input. The procedure is repeated until the desired number of frames is predicted. For label (marker) prediction a softmax layer is used. The second model, based on convolutional neural networks, aims at predicting all events directly in one single step. To evaluate their models, the authors use two benchmark datasets for action recognition and as evaluation metric, they use the accuracy of the predicted frames as mean over classes. To compare the results, they define two baselines, the first is based on a grammar method, the second on a nearest neighbor approach. In addition, they evaluate their framework using two settings: (\textit{i}) In which they use the observed labels are the ground truth annotation, (\textit{ii}) use labels provided by the decoder from a competitor – this is to show that their model is robust to errors since in real-world scenario most labels can contain errors and the prediction could be much harder. The performed experiments show the effectiveness of their model showing that the RNN model is better for short term prediction, while the CNN performs better than the RNN for longer prediction. This is since in RNN, once an activity is predicted, it is appended to the observed part and used as input to predict the next event: So, if the RNN outputs an erroneous prediction, this is likely to propagate through time.
\\

In \cite{paper13}, the goal is to predict the type of the next event conditioned on the deviation between the estimated time of the next event and the current (ground-truth) time. Their approach is focused on modeling the dependence between the marker and the time since in most real-world scenarios when an event occurs, the event marker is not available immediately, but is obtained after some time. Thus, for this reason they model the event marker exploiting the deviation between the predicted and the true time: In their hypothesis, the variation between the expected and actual time can impact the corresponding marker. Specifically, a recurrent neural network is used to learn an embedding based on the history (consisting of sequences of event type and time). The learned hidden representation is used to model the intensity function of the temporal point process (see \cite{paper23}). The predicted next time is computed exploiting the intensity function. Given the predicted and actual time, the corresponding marker is calculated using the hidden representation and the deviation (the difference between the two times) and exploiting the softmax function. To evaluate the effectiveness of their model, the authors use three real-world datasets and compare their model with several state-of-the-art approaches both for event type and time prediction.
\\

In \cite{paper15}, the authors develop two new methods to predict the evolution of the probability of the next event in asynchronous sequences, including the distributions of uncertainty. Both methods follow a common framework of generating pseudo points that describe rich multimodal time-dependent parameters for the distribution over the probability simplex. The complex evolution is captured via a Gaussian Process (GP) or function decomposition. 
The core idea of the model is to exploit the fact that the Logist Normal distribution corresponds to a multivariate random variable, and a natural way to model the evolution of a normal distribution is a Gaussian Process.
From a hidden state, the authors used a neural network to generate M-weighted pseudo points per class c. Fitting a Weighted GP to these points enables the method to model the temporal evolution. The proposed model is evaluated on several datasets and compared with two state-of-the-art methods.
\\

In \cite{paper24}, the authors propose a Deep Reinforcement Learning approach to maximize the reward returned by the environment to an agent in a setting where both the actions and feedback are asynchronous stochastic events in continuous time, modeled by Marked Temporal Point Processes. The agent's actions and the environment's feedback are modeled as sequences of pairs $(t_i,k_i)$, where $t_i$ are time instants and $k_i$ are marks. Time instants of actions (resp. feedback) are characterized by a (conditioned) intensity function $\lambda(t)$ that is the probability of observing an action (resp. feedback) in the time window $[t, t + dt)$, given the list of past events, actions and feedback, $H_t$. Moreover, the marks are characterized by a (conditioned) distribution $\mu(k|H_t)$ that is the probability of observing the mark $k$ if an action (resp. feedback) occurs at time $t$, given the list of past events $H_t$. After a time $T$, the agent receives a reward $R(T)$ which may depend on the agent’s actions and the environment’s feedback.
The time intensity and the mark distribution of the agent's actions represent its policy. It is modeled by a recurrent neural network which is trained using the list of past events $H_t$. The goal is to find the optimal agent's policy that maximizes the expected reward. 
\\

In \cite{paper34}, the authors provide a framework in which the temporal point processes are modeled by using the conditional distribution of inter-event times rather than using the intensity function. Specifically, the authors use the idea of the normalizing flows to obtain the conditional probability of the time that is modeled by using a mixture of distribution that permits sampling and computing moments in closed form. The authors model the conditional dependence of the distribution by considering the history, additional factors that can influence the distribution of the time and the sequence embedding useful when there are multiple event sequences. To encode the history, the authors use recurrent neural networks. All these factors are concatenated in a context vector used as input to learn the parameters of the mixture distribution. To show the effectiveness of the proposed framework, the authors use both synthetic and real-world datasets. The first round of experiments shows that the proposed model achieves the state-of-the-art models in time prediction task and marker-time prediction task. For the experiment, the authors show the negative log-likelihood loss. In addition, the authors show that the additional information (factors that could affect the distribution) can improve performance of the model. The last experiment is made to show that the proposed model can be also used to handle the missing data through imputation.
\\

In \cite{paper39} the authors observe that in the world many events are correlated. A single event, or a pattern of events, may contribute to cause or prevent future events. A tipical way to model sequences of correlated events is by means of Hawkes processes. This model assumes that past events can temporarily raise the probability of future events and that such excitation is (\textit{i}) positive, (\textit{ii}) additive over the past events, and (\textit{iii}) exponentially decaying with time. However, real-world patterns often violate these assumptions. Indeed, (\textit{i}) is violated if one event inhibits another rather than exciting it, (\textit{ii}) is violated when the combined effect of past events is not additive and (\textit{iii}) is violated when, for example, a past event has a delayed effect, so that the effect starts when the event occurs and increases sharply before decaying. A basic model of event streams assumes that an event of type $k$ occurs in the infinitesimally time interval $[t; t + dt)$ with probability $\lambda_k(t)dt$. The paper generalizes the Hawkes process and overcomes its limits by determining the event intensities from the hidden state of a continuous-time LSTM that has consumed the stream of past events. In this way, they are able to influence the future in complex and realistic ways. This model has been named \emph{neural Hawkes process}.
\\

In \cite{paper40}, the authors model the background by a RNN with its units aligned with time series indexes, while the history effect is modelled by another RNN whose units are aligned with asynchronous events to capture the long-range dynamics. The whole model with event type and timestamp prediction output layers can be trained end-to-end.
Time series and event sequences are fed into two RNNs (LSTM) connected to an embedding mapping layer that fuses the information from two LSTMs. Then three prediction layers are used to output the predicted type, sub-type of events, and the associated timestamp. Cross-entropy with the time penalty loss and square loss are respectively used for event type and timestamp prediction.
A comparative analysis with several state-of-the-art approaches on a real-world dataset is performed to show the effectiveness of the proposed model.
\\

In \cite{paper41}, the authors provide a Long-Short Term Memory Model to estimate when users will return to a site and their future listening behaviour. The authors try to solve the problem of Just-In-Time recommendation, that is, by recommending the right items at the right time.
Learning recurrent users' interests fits nicely into the framework of the recurrent network. The past session's actions for a user are fed to an LSTM and let him figure out how to predict the actions for the next session. The negative Poisson log-likelihood is used as a training objective. With the recurrent visiting patterns and action tasks, the model solves them jointly. The parameters in LSTM and all the embedding matrices are then trained by minimizing the combined objective. The authors show the capability of the proposed model compare it with several competitors and using two real-world datasets.
\\

In \cite{paper42}, the authors propose a novel deep point process to make the model generalize well across tasks, integrating multiple inputs flexibly and maintaining model interpretability by uncovering causal influence between events. To this end, the proposed deep point process consists of three components: First, one temporal recurrent neural network capturing event interaction dynamics; second, one synchronous RNN updating the exogenous intensity based on time series; and third, an attention mechanism introduced to uncover influence strengths among events. An attention mechanism for the point process is devised to improve the model interpretability. With the help of the attention mechanism, the event dependence encoder can choose which parts of a long sequence to focus on for current prediction instead of encoding all the historical events into one hidden variable. To show the effectiveness of the proposed model, the authors use several real-world datasets and provided a comparative analysis with several competitors.
\\

\cite{paper43} deals with the problem of predicting the diffusion dynamics of cascades.
A cascade models how information diffuses across a social network platform. In other words, the proposed approach tries to predict how information gradually \emph{infects}
the nodes of a social network. In this context, a node is infected if it takes part in the cascade process (i.e., re-shares a content). More formally, a cascade $c$ is a sequence of tuples $(v_i, t_i)$, where $v_i$ are the identifiers of the infected nodes, $t_i$ are the infection times, and $i \in [1..n]$. Given a cascade $c$, the problem is to predict the next infected node $v_{n+1}$ and the next infection time $t_{n+1}$.
In the paper a first vanilla LSTM model is proposed. Each node $v_i$ is codified with an embedding $x_{v_i}$ of fixed size computed during the training process and for each infection time $t_i$ the inter-infection duration $d_{t_i}=t_i-t_{i- 1}$ is computed. The recurrent neural network processes the sequence of the node embeddings and the inter-infection durations $(x_{v_i}, d_{t_i})$. The prediction of the next infected node and the corresponding infection time is performed leveraging the hidden state value $h_n$ of the LSTM network. A multi-class classifier, whose input is $h_n$, with a softmax output layer is used to predict the next infected node. The next infection time is predicted by computing a conditional intensity function $f_n (\cdot)$. A training set of cascades is used to learn the node embeddings, the LSTM parameters and the classifier parameters. Then, the test cascade is processed by the trained model in order to compute its last hidden state end predict the next infected node and the next infection time. The \emph{DeepDiffuse} model is an improved version of this architecture that introduces an \emph{attention mechanism}. The idea is to compute the output of the LSTM not just by the last hidden state but as a weighted summation of all the hidden states. Finally, the authors proposed some enhancements to this model in order to provide more flexibility to its attention mechanism.
\\

In \cite{paper45}, the authors describe a novel activity prediction method for smart houses based on the seminal probabilistic method named Marked Temporal Point Process Prediction. The authors applied MTPPs to a freely available dataset in the intelligent house community provided by the CASAS project. The model prediction is based on the RMTPP’s architecture \cite{paper3}. The model differs substantially since time series and event sequences are predicted separately. Specifically, the model receives different timestamps into an LSTM, and the sequence of activities is embedded and fed into a separate LSTM. The recurrent layer learns a representation that summarises the non-linear dependency over the previous events. The hidden unit of RNN enables the model to learn a unified representation of the dependency over history. In terms of the evaluation protocol, the authors used a real-wold dataset to assess the performance of the proposed method comparing it with three state-of-the-art methods in both event type and time prediction tasks.
\\

An improved version of \cite{paper45} is provided in \cite{paper93} in which the proposed model is based not only on RMTPP, but also on ERPP \cite{paper40} architecture. The results show that the two architectures achieve very similar performances on all datasets: however, since in the activity daily living prediction the activity time is strongly correlated with the time in which it is performed, the strategy of learning them separately in ERPP is better.
\\

\cite{paper46} deals with a special case of Marked Temporal Point Processes, where the event sequence is an event propagation process over a directed graph and the markers denote the nodes of the graph. For example, a retweeting sequence in social network where the markers denote users.
Given a collection of event propagation sequences, conventional point process models consider only the event history, that is embed into a vector (usually the hidden state of a recurrent neural network), ignoring the latent graph structure. The authors propose the Graph Biased Temporal Point Process (GBTPP) model that leverages the structural information of the graph, provided by a graph representation learning technique. It allows to model the direct influence between nodes and indirect influence of the event history. In more detail, conventional TPP models solve the event propagation problem by processing sequences of pairs $(v_i, t_i)$, while the GBTPP model leverages the structural information of the graph by processing tuples $(v_i, t_i, y_i)$, where $y_i$ is the node embedding vector obtained by a graph representation learning method.
\\

The authors in \cite{paper54} propose Topo-LSTM, a novel topological recurrent neural network based on dynamic DAGs, for the diffusion prediction task. In particular, they focus on cascades, trying to estimate the probability of an inactive node being activated next. Cascades are understudied in the literature, with traditional methods requiring manual efforts and extensive domain knowledge to perform the feature engineering process and deep learning methods frequently underexploring cascade topologies. To model the diffusion topologies, the authors develop a DAG-structured RNN, which takes dynamic DAGs as inputs and generates a topology-aware embedding for each node in the DAGs as outputs. They specifically adapt the memory cell design of the standard LSTM to these changes, extending it to include different input types and multiple inputs for each type. Topo-LSTM is tested on three real-world datasets and against several state-of-the-art competitors.
\\

In \cite{paper55}, the authors propose an attention-based RNN to capture the cross-dependence in cascade. Furthermore, they introduce a coverage strategy to combat the misallocation of attention caused by the memorylessness of traditional attention mechanisms. For modelling cascade dynamics, the authors propose to construct both source and target sequence from the observed cascade, restricting that the (k + 1)-th resharing behaviour is the output of the k-th resharing behaviour. Additionally, they propose a coverage mechanism to adjust the misallocation of attention, leading the alignments to reflect the true structure of propagation better. The memoryless characteristic in the attention mechanism causes the misallocation of attention.
The proposed method is tested on a synthetic dataset and against several state-of-the-art approaches.
\\

In \cite{paper56} methods for leveraging temporal information for event sequence prediction are proposed. Based on our intuition about how humans tokenize periods and previous work on the contextual representation of words, the authors proposed two methods for time-dependent event representation. The proposed method transforms a regular event embedding with learned time masking and form time-event joint embedding based on learned soft one-hot encoding. They also introduced two methods for using the next duration as a way of regularization for training a sequence prediction model. The goal here is to predict the next event, which can help to learn by introducing an additional loss component based on the prediction of the next event duration. The duration prediction of the next event at the step is computed from a linear transformation of the recurrent layer. The authors perform experiments on several real-world datasets using different competitors.
\\

In \cite{paper58} is addressed the problem of next event mark and time prediction by proposing to model marked temporal dynamics with a function that captures their interdependence.
 In fact, existing methods either predict only the mark or the time of the next event, or both separately.
 For these reasons, the authors propose an RNN-based method (RNN-TD) that uses a mark-specific intensity function to model the occurrence time of events with different marks. Furthermore, the RNN embeds sequential data characteristics, simplifying the modeling of dependency on historical events via a deep structure. 
 The proposed model is tested on two large-scale real-world datasets with several state-of-the-art methods. RNN-TD outperforms them in the prediction of marks and times.
 \\

In \cite{paper64} is applied well known methodologies for Temporal Point Processes to analyze sequences of object reviews. The proposed approach has been presented in the domain of \emph{conversational recommender systems} but can be applied to other domains.
It exploits a RNN-based model to process a sequence of reviews related to an object and, capturing the changes of user preferences over time, predict the content and the time of the next review. In detail, each review is transformed into a \emph{Bag of Words} (\emph{BoW}) representation, a vector modeling its content. The sequence of BoWs and timestamps related to an object is processed by a special type of LSTM model whose hidden state is used to derive the next BoW and the next timestamp.
\\

The authors in \cite{paper66} propose a model for predicting the next-event marker and time-to-event. It takes as input a prefix, i.e., a sequence of events, and models the sequence as a compact, low-dimensional embedding that is then passed to a temporal point process for estimating the probability density. They identify the modeling of the solely temporal dependence between events as a common limitation of the RMTPP approach. For this reason, they design a novel architecture that employs a hierarchical attention mechanism to model the importance of each entity type in the sequence. The proposed architecture produces more informative sequence embeddings, then fed to a temporal point process to estimate the probability density on event markers and time duration. Experimental results on two social media datasets and one financial transaction dataset prove the capability of Multi Relation Structure-RMTPP to capture subtle nuances of sequential data better.
\\

In \cite{paper70} a neural memory network-based framework for future action sequence forecasting is proposed. This is a challenging task where we have to consider short-term, within-sequence relationships and relationships between sequences to understand how sequences of actions evolve. The authors introduce neural memory networks to our modelling scheme to capture these relationships effectively. The inputs are encoded using a ResNet50 and a categorical function Fcat, then are temporally mapped using LSTMs. The authors proposed two memory modules to model the long-term relationships of the individual modalities. These individual memory outputs are concatenated and passed through another LSTM and a fully-connected layer, generating the future action sequence. Two real-world datasets are used to show the effectiveness of the proposed model against state-of-the-art competitors.
\\

In \cite{paper71}, given a sequence of labeled data, ($\{(x_1, k_1), ...(x_n, k_n)\}$ where $x_i$ is the time and $k_i$ is the marker) and unlabeled data, ($\{x_1, ..., x_n\}$) the authors want to predict the next event marker and time exploiting both labeled and unlabeled data. Indeed, in real world scenarios only a limited amount of labeled data is available: Specifically, the authors assume that only the marker information is missing. The proposed model consists of two branches: The first, i.e., the labeled branch, allows the model to learn the relationship between the markers and the times, the second, i.e., the unlabeled, is used to learn a representation for the time. The labeled branch is composed by recurrent neural network (LSTM) and provides as output an encoding that summarizes the contribution from the time and the marker (f=RNN(x, k)). The second branch is an RNN-based encoder decoder. The unlabeled path provides the embedding (h=ENC(x)) of the time that is fused with the embedding provided by the labeled path ($f+\lambda * h$). The fused representation is fed into a multi-layer perceptron network to predict the next time and marker. The proposed model is trained by minimizing the following loss: $L_{time} + L_{marker}+ L_{recon}$. The proposed approach is compared with a native supervised MTPP model which follows a similar architecture of the proposed model without the unsupervised (unlabeled) branch. Experiments have been performed on the Retweet dataset and six protocols containing varying labeled data are used. The results, in terms of average precision, macro-F1 and micro-F1, show that the proposed model improves the performances of the supervised model. In addition, the authors also provide an experimentation to analyze the effect of the weight parameter ($\lambda$) used for the fusion representation to understand the effect of the fusion of the supervised and unsupervised representations.
\\

\begin{figure}[h]
    \centering

    \begin{subfigure}[b]{0.47\textwidth}
         \centering
         \includegraphics[width=\textwidth]{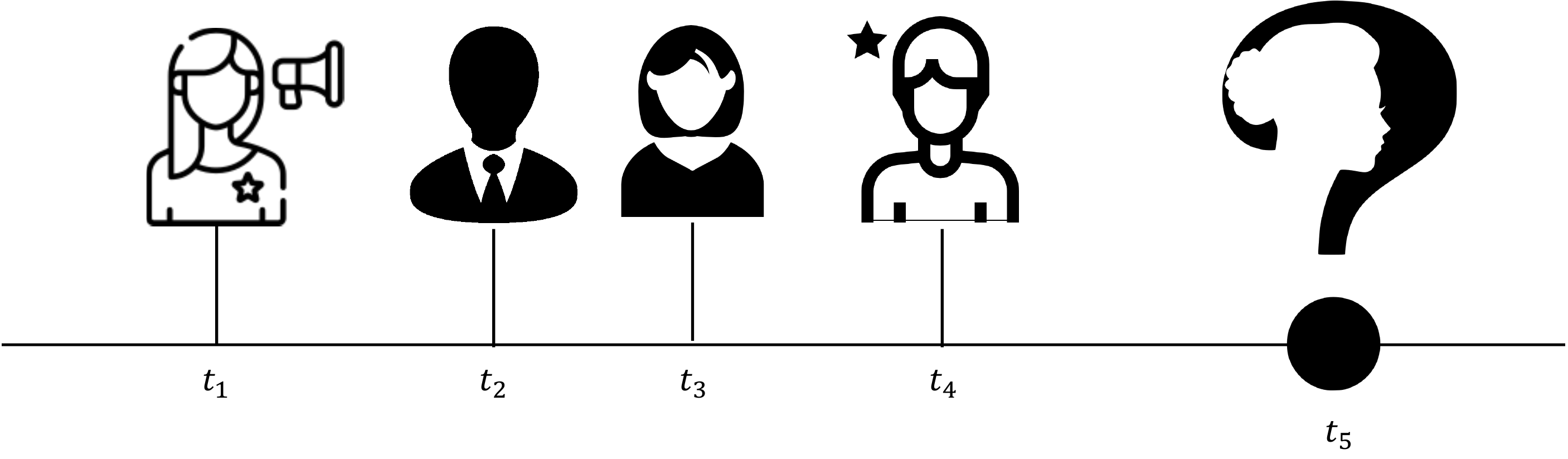}
     \end{subfigure}
     \qquad\qquad
     \begin{subfigure}[b]{0.43\textwidth}
         \centering
         \includegraphics[width=\textwidth]{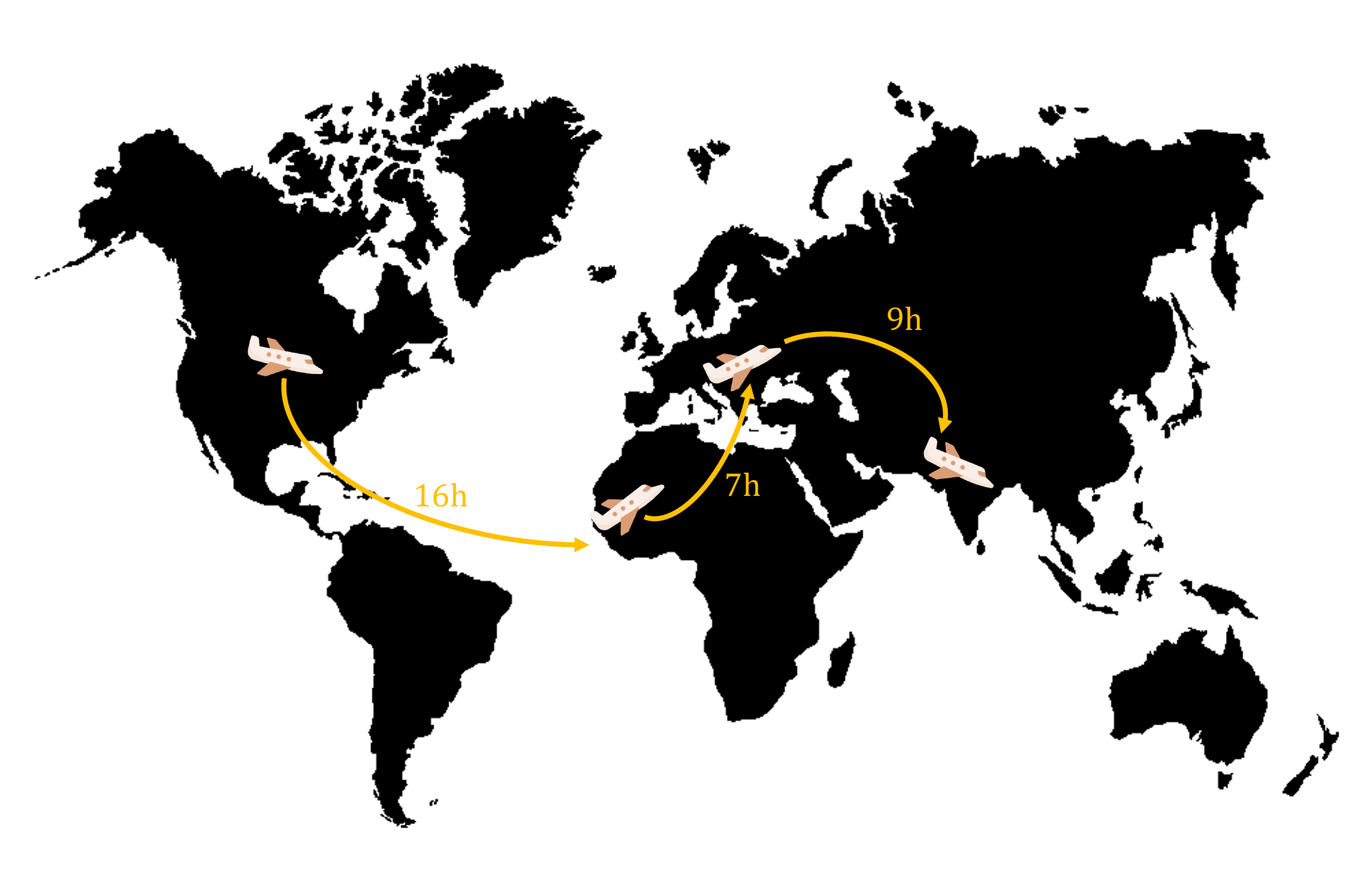}
     \end{subfigure}
    \caption{Two examples of Marked Temporal Point Process. (Left) On a platform, a user of a certain type (marker) posted at time $t_1$ a content. This event influences the other events, i.e., other users that share the principal post. The goal is to predict which user will re-post the content in the future and at which time. (Right) The event is the airplane flight and the aim is to predict which will be the next destination of the airplane and how much time will be spent to reach the destination.}
    \label{fig:paper72}
\end{figure}

To predict future occurrences of events in the form illustrated in Fig. \ref{fig:paper72}, that is their types and their timing, a number of Marked Temporal Point Process models have been designed. Most of them make use of recurrent neural networks and are fundamentally based on the Maximum Likelihood Estimation for the training. However, their likelihood function is difficult to estimate because part of it is a definite integral whose computation in general is intractable. The existing approaches overcome this problem by (\textit{i}) approximating the likelihood with Monte Carlo sampling or (\textit{ii}) making use of integrable functions. In \cite{paper72} the authors proposes an alternative framework to train MTPP models, whose likelihood function is intractable, called INITIATOR (noIse-coNtrastIve estimaTIon for mArked Teemporal pOint pRocess). It is based on the Noise-Contrastive Estimation (NCE) method, where parameters are learned by solving a binary classification problem. For each event, a 'noise sample' is generated and the classifier has to distinguish between true samples and noise samples. Training the classifier implies computing the optimal parameters of the Conditional Intensity Function.
\\

Most existing MTPP models consider only the complete observation scenario, i.e., the event sequence being modeled is completely observed with no missing events.
This is an ideal setting that sometimes is not applicable in real-world scenarios.
In \cite{paper73}, the authors present a novel modeling framework for point processes called IMTPP (Intermittently-observed Marked Temporal Point Processes) able to learn the dynamics of both observed and missing events.
In this framework, the missing events are represented as latent random variables, which together with the previously observed events, seed the generative processes of the subsequent observed and missing events.
The IMTPP architecture includes three generative models, based on recurrent neural networks: MTPP for observed events, prior MTPP for missing events, and posterior MTPP for missing events.
It is worth noting that the sequence of training events that are provided as input to IMTPP consists of only the observed events, while the missing event sequence is modeled through latent random variables, which, along with the previously observed events, drive a unified generative model for the complete event sequence (observed and missing events) .
\\

In \cite{paper77} ProActive is described, i.e., a neural marked temporal point process framework for modelling the continuous-time distribution of actions in an activity sequence while simultaneously addressing three high-impact problems – next action prediction, sequence-goal prediction, and end-to-end sequence generation. Specifically, the model utilizes a self-attention module with temporal normalizing flows to model the influence and the inter-arrival times between actions in a sequence. Moreover, the model detects sequence goals early for time-sensitive prediction via a constrained margin-based optimization procedure. This, in turn, allows ProActive to predict the sequence goal using a limited number of actions. The proposed model is tested on three real-world datasets against state-of-the-art approaches.
\\

In \cite{paper79}, the authors define and introduce a Temporal Hierarchical Recurrent Neural Network (THRNN): A joint model for inter-session and intra-session recommendations and return-time prediction. It consists of three major components, a module for the sequence of sessions of representations, a module for sequences of user-item interactions within each session, and a module for the time interval between the sessions. The first two modules consist of a Hierarchical Recurrent Neural Network (HRNN) architecture, and the third module is given by a Point Process which shares representations with the HRNN. A tuning mechanism in the training process allows the time model to modulate the focus on short, medium or long-time prediction. The mechanism adds a control parameter in the loss function that allows us to control the importance of temporal information at training time. Two real-world datasets are used to test the capability of the proposed model against state-of-the-art methods.
\\

\subsubsection{Generative Models (VAE and GANs)}

\begin{figure}[h]
    \centering
    \includegraphics[width=.7\textwidth]{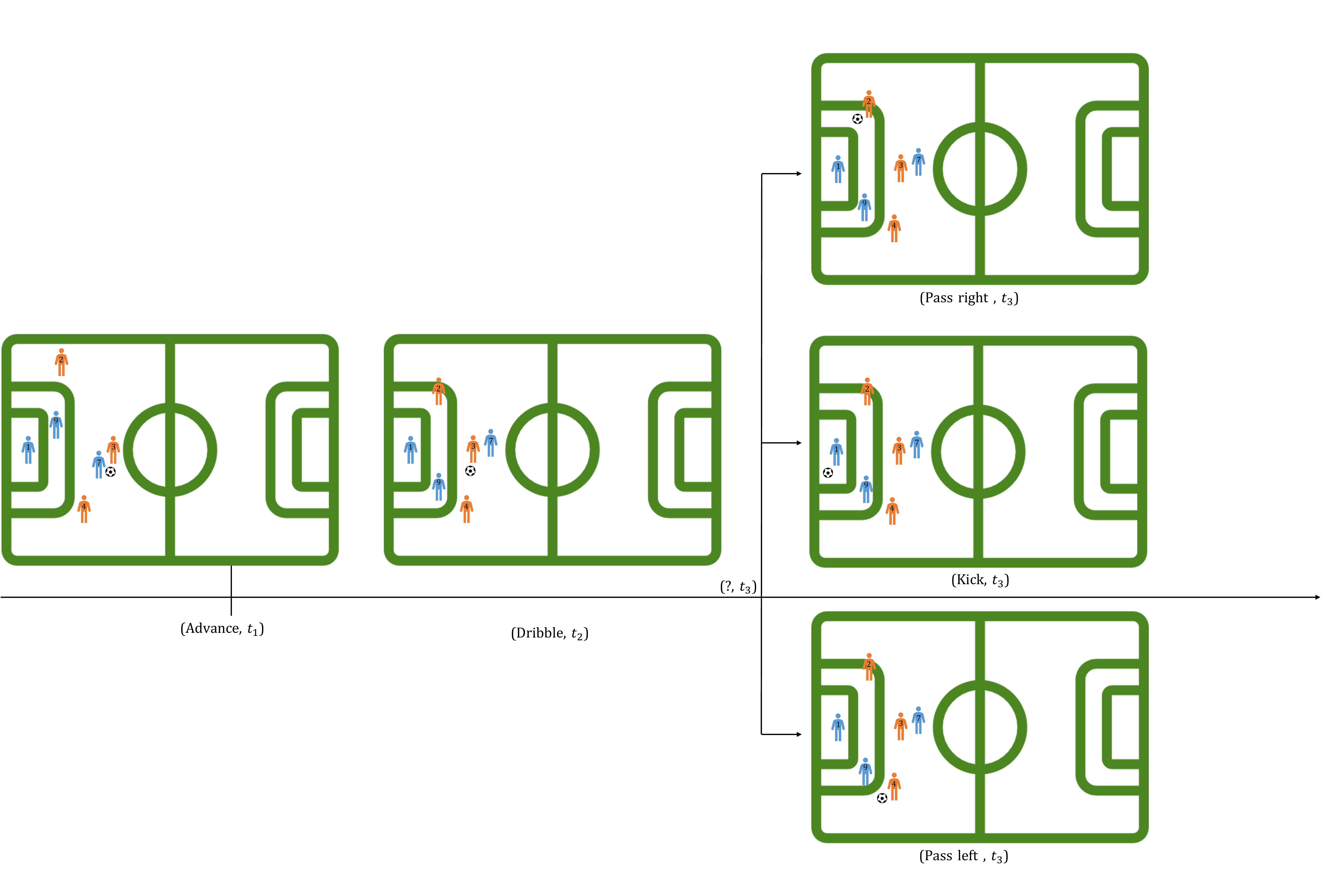}
    \caption{Evolution of a marked temporal point process. The game evolves through the actions \textsl{advance} (at time $t_1$) and then \textsl{dribble} (at time $t_2$). Three further scenarios are possible at time $t_3$, which characterize the evolution with high uncertainty.
    }
    \label{fig:paper2}
\end{figure}

In \cite{paper2} the goal is to predict the next action (event type) and the time at which the action occurs (the authors use the term “inter-arrival time” to indicate the difference between the starting time of the previous action and the current action). The authors propose a model based on Variational Autoencoder: Specifically, given a sequence of actions, in which each action is composed by the marker and the inter-arrival, the model produce two probability distribution: One over the possible action markers and another over the inter-arrival time. The encoder is instantiated as a recurrent neural network (an LSTM is used). The input of the recurrent neural network is the concatenation of the marker and inter-arrival representations. Given this new representation, the model is composed by two networks (LSTM): One is used to estimate the parameter of the posterior distribution and another for the parameters of the prior distribution (unlike the original VAE, the prior is not fixed but it is learned). During the training, a latent variable is sampled by using the posterior distribution and it is fed to the action decoder and the time decoder for generating distributions over the action marker and inter-arrival time respectively. At test time, given the prior distribution it is possible to predict the next marker and the inter-arrival time (in this way, it is possible to consider the uncertainty since the prior distribution can be used to sample different possible evolution. Figure \ref{fig:paper2} shows an example history of events describing a sequence of actions. We can see that, on the last known event, a model must be able to reason on many different possible evolutions). The authors use the log-likelihood, the accuracy for action prediction and the mean absolute error for the time. For the comparison, the authors use one state-of-the-art approach and propose two baselines, i.e., APP-LSTM in which there is no stochastic latent variable that allows capturing the uncertainty, and APP-VAE w/o learned prior in which the prior is fixed to the standard normal distribution. The experiments show that the proposed model outperforms the baselines and the state-of-the-art approach.
\\

In \cite{paper5} the authors show an effective way to adapt the framework of Generative Adversarial Networks to the case of sequential modeling. Specifically, they devise a method that given a sequence of events is able to effectively predict the next event type and its timestamp. The framework consists of a simple network architecture based on recurrent neural networks (LSTM blocks) and a naive feature encoding - i.e., one-hot encoding of event labels. Experimental analysis show that the adversarial training devised in this work is able to provide a better accuracy w.r.t. other non-adversarial competitors at the same training cost, and it is also able to leverage the prefix length by showing a monotonic increase of performance when the length increases. 
\\

In \cite{paper6} the problem is to predict how a partial sequence of time-stamped events (process logs) will continue till termination of the process, also predicting timestamps. The methodology adopts a GAN where the Generator is an Encoder-Decoder based on RNNs, and the Discriminator is a LSTM. After training, the Generator is used as predictor. The novelty w.r.t. previous work (ref. \cite{paper5} by the same authors) is mainly in predicting whole sequences (thanks to the encode-decode architecture) rather than reiterating single event predictions. 
The evaluation is based on (1) edit distance of the produced sequences of events, and (2) the total duration of predicted sequences (= time left to complete the process). Results show better performances w.r.t. \cite{paper5}, especially on long sequences. An improved version of \cite{paper6} is provided in \cite{paper9}.

In \cite{paper16}, the authors propose a variational autoencoder trained to fit the conditional intensity function, i.e., the function that returns the expected rate of events at an instant of time, given a series of past events. For each event $e$, its type and its time interval - the time elapsed between the previous event and $e$ - are encoded using a recurrent model. The model leverages past events to compute a distribution over a latent space. Then, a sample of this space is decoded into two probability distributions: One over the possible event types and another over the time intervals for the next event. They are used to predict the type of the next event and its time interval and to compute the loss w.r.t. the ground truth given by the actual current event. In detail, for each event, an embedding for its type and an embedding for its time interval are computed. The two embeddings are concatenated in order to obtain a composite embedding of the event. The sequence of these embeddings is processed by an LSTM model whose hidden state is used to compute the probability distribution over the latent space.
\\

In \cite{paper17}, the authors provide a transformer-based model that incorporates information from video, dialogue, and commonsense knowledge, each of which is necessary for Video-and-Language Event Prediction (VLEP) - given a video with aligned dialogue and two future events, an Artificial Intelligence (AI) system is required to predict which event is more likely to happen. The model encodes each video frame's appearance and motion features. For appearance, the model extract 2048D feature vectors from the pre-trained ResNet-152. For motion, 2048D feature vectors are extracted from pre-trained ResNeXt-101. The model performs L2- normalization and concatenates the features as the video representation. The model uses the text's contextualized text features from the RoBERTa. The concatenation of dialogue and future event candidates as input to the transformer layers, unique tokens such as CLS is also added in this process.
\\

In \cite{paper18}, the authors propose a VAE-based model to predict the next event type and the time at which the event occurs. They use transformer structures for the inference and generative networks. The inference network aims at predicting the conditional intensity function using the learned latent variable, while the generative model aims at reproducing the input sequences of events to help the learning of the latent variable. An encoded representation consisting of an embedding layer to encode event types and a temporal embedding is fed in the transformer. The output of the transformer network is used to produce the mean and variance of the latent variable used to obtain the intensity function. The authors propose two versions of the intensity function: In the first there is a linearly decaying w.r.t. time, while the second presents an exponentially decaying. Then the intensity function is used to predict the next event type and time using numerical approximation. To evaluate our model, the authors use both real-world and synthetic dataset and compare their model with two Hawkes processes and two state-of-the-art neural networks. The first experiment performed wants to show the performance of their model compared to the competitor in the event type and time prediction; for that experiment they use F1-score for the type and RMSE for the time. In addition, since their model can compute the conditional intensity function, they compare the predicted conditional function with the true for the synthetic dataset showing that their model outperforms state-of-the-art methods. The authors also show that the learned latent variables can represent the history and next event type: They conduct a singular value decomposition, and the results show that the latent variables are automatically separated according to the event type showing that the model is able to learn the different representations for each event. The last experiment is performed to show that their model can generate the most diverse event types.
\\

In \cite{paper20}, the authors propose a double Wasserstein generative adversarial network (WGAN) for generating types of events with their times, where two WGAN models, one for event times and another for their types, are trained simultaneously. Specifically, the authors use the first WGAN to generate the time starting from an input noise generated by a Poisson distribution. Then the sequence times and a random noise are fed in another WGAN to generate a sequence of type vectors. This WGAN is conditional since the authors want to model the inter-dependence between the time and the event type. In addition, the Gumbel-Softmax layer and type embedding layers are utilized to handle types of events – they are categorical data. Both generator and discriminator are composed by recurrent neural networks. To evaluate the proposed model, the authors use both synthetic and real-world datasets. Since the model predicts both time and type event, different metrics are used: Specifically, for time prediction task the intensity deviation metric is used, while for type prediction, information retrieval metrics (i.e., weighted precision, Top-1, Hits@K scores) are used together with the intensity deviation for event type metric. The intensity deviation is defined as the difference between the average generated intensity and the average real intensity divided by the magnitude of real intensity. To show the superiority of their model, the authors decide to compare their model with two functional models and with the same model in which there is no dependency between the time and the type. The experiments show that the proposed model achieves comparable results with the baselines.
\\

The framework proposed in \cite{paper44} leverages what is called \emph{social homophily} among users, i.e., similarities in their traits or interests, and temporal dynamics to predict the next infected user. The temporal dynamics are influenced by two factors: The order of the users in the cascade and the effects of their popularity. The authors propose an architecture based on a Variational Autoencoder to learn: (\textit{i}) A latent representation $h_u$ for each user $u$ that captures his/her social homophily w.r.t. other users; (\textit{ii}) two types of embeddings modeling the temporal and popularity factors of a cascade and that are merged by an attention fusion strategy in order to compute the receiving capability $v^r_u$ of each user $u$, i.e., the his/her attitude to be the next infected user. After the training phase, the likelihood of influencing user $v_j$, given a test cascade $C$, is given by $p(v_j|C)=\sigma(h_j^T \cdot v^r_j)$.
\\

In \cite{paper68} HGANPP, an adversarial-based marked temporal point process framework capturing hashtag popularity dynamics is proposed. The authors' goal is to design a novel architecture that takes into account not only the intensity aspect of a classical point process but also side information regarding hashtag micro-dynamics. By proposing these advancements, the authors want to tackle the significant issue of state-of-the-art models providing ineffective forecasting. This novel framework employs a linear semi- auto-regressive model for event-type generation and couples the event type and time aspects. It also considers two different phenomena influencing hashtag popularity: (\textit{i}) Hashtag-tweet reinforcement, the phenomenon for which hashtag and tweet popularity mutually reinforce each other, and (\textit{ii}) inter-hashtag competition since hashtags do not simultaneously flourish at the same time; indeed, they rise/fall at the cost/advantage of other hashtags. HGANPP considers at each time step $t_i$ the retweet count of a hashtag and also introduces a regularization loss based on the popularity rank of the hashtag w.r.t. other hashtags. Experimental analysis on seven datasets and several competitors show how HGANPP significantly outperforms current state-of-the-art models. It also proves to be more effective in forecasting tasks. Furthermore, it is independent of the modality through which a hashtag becomes popular, remarking its ability to learn beyond a fixed functional form.
\\

\subsubsection{Neural Networks}
In \cite{paper7} the authors provide a comparison of existing methods – more exactly, 3 Deep Neural Network (DNN) architectures (Multi-Layer Perceptron (MLP), RNN, CNN), each implemented with 5 different methods for encoding context information ((binary, ordinal, onehot, hash and word2vec). Tests are performed on 5 real datasets in business process management. The authors show that LSTM (RNN) is the best architecture, hashing is the best encoding, class imbalance has significant impact.
\\

In \cite{paper8} a framework for suffix prediction in the context of Process Mining using Deep Learning is proposed. The authors initially highlight how suffix prediction is the most challenging task in Process Mining. Then they proceed to stress reproducibility issues since the current literature evaluates the performance of DL models on small datasets and that comparing different paper results is hardly doable because of different pre-processing and evaluation strategies. This framework introduces a standard procedure based on embedding, various families of sequence modeling, and generator modules. Finally, they evaluate this framework using different sequential modeling techniques on several datasets. They prove that there is no clear winner and that, with increasing prefix length, the performance highly fluctuates, thus posing serious concerns about the standard way to measure performance through simple averages.
\\

Given text sequences, \cite{paper10} aims to predict the next event in the form of grammar units (verb, verb+subject, etc.). The last event in the sequence is compared to all possible candidate events, and the most similar one is selected as prediction. Similarity is given by a neural network, based on various possible representations of the two input events to compare: Latent semantic indexing, word2vec of single words, word2vec of verb + subject and context (which is the best performer). Testing is based on a novel procedure, where the predictor is given 5 candidates to choose from. 
\\

In \cite{paper11}, the authors propose Attentive Neural Point Processes (ANPP) to solve the problem of predicting the next event given a sequence of events happened in the past. The main contribution of this model is to adapt the self-attention mechanism to model the time-level pairwise dependency of each historical event. Basically, each marker (type of event) is embedded into a latent space, and it is further weighted by a self-attention layer that takes into consideration all the other markers in the sequence. Remarkably, time is not treated as a scalar value (thus regressed), but it is bucketed in a discrete space defined as inter-event duration bucket embedding. The neural architecture is optimized by means of a negative log-likelihood that weights the losses of markers and times. ANPP proves competitive results w.r.t. other point process approaches across different datasets and different sequence lengths. Notably, authors show anecdotal evidence of the explanation capabilities of the learned attention coefficients allowing to highlights the users' interests.
\\

In \cite{paper19}, the authors build several neural network models to predict the CDF of the time to event (TTE) for a specific event type, e.g., a vehicle stop, over video data. The video nature of data is dealt with by a 3D-ResNet 34, thus reducing videos to fixed-length embeddings. Both solutions for continuous and discrete time representations are considered, including multi-class classification and heatmaps over binned values, regression, parametric distributions (Gaussian, Weibull) and an hybrid heatmap where single probabilities are represented through Gaussians and they are mixed to learnable weights.
\\

In \cite{paper30} the authors focus on modeling temporal interaction networks, i.e., user-item networks where each edge is marked with a timestamp. They identify a significant limitation in current approaches based on recurrent neural networks: They do not consider the structural information coming from the network topology.
For this reason, they introduce two novel deep encoders named Topological Feature Encoder (TFE) and Attentive Shift Encoder (ASE) to characterize the stable intentions of users and items and the long-range time dependency of the observed interactions. The output of the blocks mentioned above is given as input to an intensity function similar to the one used in Multivariate Hawkes Processes (MHP). Finally, the authors benchmark the framework on two different tasks: The item prediction task given a user $u$ and a time $t$, and the time prediction task given a user $u$ and an item $i$. Experimental analysis against several competitors confirms the proposed model's superior performance on item and time prediction tasks.
\\

The authors in \cite{paper35} want to address the problem of predicting the next type event and its time, given its history. They propose a self-attentive Hawkes process where self-attention is used to summarize the influence of history events used to predict the next event. First, since self-attention utilizes positional encoding to retrieve the order information into a sequence, to consider time intervals of subsequent events, the authors propose a new positional encoding in which time intervals are translated into phase shifts of sinusoidal functions. This encoding along with the embedding of the event type are used as input for the self-attention module (in particular, the self-attention module takes as input the history, so for each event in the history we have the concatenation of the two encoding). The self-attention module produces as output an hidden representation that summarizes the influence of all previous events. The hidden vector is used to compute the intensity function. To evaluate the ability of their model, the authors use both synthetic and real-world datasets and compare their model with different state-of-the-art approaches. First, to show that their model can better approximate the intensity function, the authors use the synthetic dataset where the true intensity is known. The authors use the negative log-likelihood to show that their model can model a specific event sequence: In fact, w.r.t. the competitors, their model achieves better results. For the prediction task, the authors use the F1-score and the root mean square error for the event type and the time, respectively. The results show that the proposed model outperforms the competitors. In addition, the authors demonstrate that their model is more interpretable than the RNN-based model because the learnt attention weights reveal contributions of one event type to the happening of another type.
\\

In \cite{paper38}, the authors define a framework based on transformer models to predict, given a partial sequence of events, what type of event will happen next and its time. Specifically, the authors decide to use transformer models to capture short-term and long-term dependencies between the events in the sequence. Transformer model takes as input the concatenation of the encoding of the time and the marker and returns as output the hidden representations of all the events in the input. Given that representation, it is possible to define different intensity functions for different event types. Each intensity function is composed of three terms: The current influence, the history influence and the base influence. Given the intensity function, the authors can model the next time and event type. The authors also show that their approach can be generalized to consider more complicated data, such as event sequences on graphs. For the evaluation, the authors consider per-event log-likelihood; for event prediction, they consider accuracy for the marker and root mean square error for the time. The authors compare their model with several state-of-the-art approaches and on several real-world scenarios showing that the proposed approach achieves competitors’ performance.
\\

In \cite{paper57}, the authors describe the LSHP, composed of the multi-dimensional Hawkes process that captures the “mutual influence” of different actions within a period and the one-dimensional Hawkes process that captures the “self-influence” of actions of the same type across different periods. Intuitively, mutual influence reflects the transitional patterns among different actions in close temporal proximity, and self-influence characterizes repetitive patterns of the same type of actions over a longer period. Because the mixture of these two types of behaviour dynamics behind an observed action sequence is latent, the proposed method model it in a probabilistic manner and estimates model parameters via a maximum likelihood estimator. The authors conduct experiments on two real-world datasets compared to several state-of-the-art approaches.
\\

\cite{paper88} deals with the problem of making predictions in the context of Multivariate Hawkes Processes and proposes a simple non parametric technique to estimate the intensity functions related to pairs of nodes. Let us suppose to have $d$ nodes (e.g., the users of a social network).
For each node $i$ and instant of time $t$ let us consider $N_t^i$ the number of events associated to $i$ up to $t$. 
Hawkes processes are modeled by an autoregressive structure of the intensity functions in order to capture self-excitation and cross-excitation of nodes. These phenomenona are observed, for instance, in social networks.
In this case the intensity function of a node $i$ can be defined as $\lambda_t^i = \mu_i + \sum_{j=1}^d \int_0^t \phi^{ij}(t-t')dN_{t'}^j$, where $\mu_i$ is a constant and $\phi^{ij}(\cdot)$ 
is an integrable function, called kernel, modeling the impact of an action of node $j$ on the activity of node $i$.
Note that when all kernels are zero, the process is a simple homogeneous multivariate Poisson process. Most of the literature uses parametric approaches for estimating the kernels. In this paper the authors propose a more direct approach. Instead of trying to estimate the kernels they focus on the direct estimation of their integrals.
\\

\subsection{Traditional Statistical Methods}
In \cite{paper14}, the authors propose a model to predict the final number of resharings of a social-media post. They build a statistical model upon self-exciting point processes with three main advantages: (\textit{i}) It requires no training and has real-time inference times, (\textit{ii}) it is accurate, and (\textit{iii}) it also exhibits an interpretable "infectiousness" parameter expressing whether the information cascade is in a super-critical or sub-critical state. Self-exciting point processes (SE-PP) are counting processes that count the number of instances over time. The advantage over standard counting process - such as homogenous Poisson - is that SE-PP do not assume constant intensity over time, instead they model the influence of all previous instances on the current evolving behavior. Authors contributes by introducing a new technique based on Galton-Watson trees to model the temporal patterns of social-media information spreading. The model is scalable, accurate, explainable and it also requires minimal amount of information since it does not require full knowledge of the social network but only the out-degrees of the nodes involved in each cascade.
\\

In \cite{paper53} a joint model for link prediction and “retweet” generation in a social network is developed. In particular (the most novel part) the link generation between nodes $u$ and $s$ is modeled as an Hawkes process triggered by $u$ retweeting messages of $s$ (the source of information arrived to $u$ indirectly, through common neighbors). The retweet process, as usual, follows the (dynamic) links structure and another Hawkes process. An efficient (incremental) simulation algorithm is given, which alternates the two steps (generate tweets - generate links), and also an efficient parameter estimation procedure is provided. Results show better performances on link prediction w.r.t. state-of-the-art approaches.
\\

In \cite{paper65} the authors tackle the problem of predicting the popularity of social media content. Similar to previous works, they leverage self-exciting processes to model how the occurence of an event stimulates the likelihood of another event nearby in time. They also exploit the number of followers of each social media user to characterize the extent to which each user is able to propagate the content to other peers. Although previous literature highlighting similar elements, authors identify a major limitation in that those works do not take into account the mutual influence between different activity types such as Retweet, Reply and Like. For this reason, they adapt Ogata's Epidemic-Type Aftershock Sequences models - also known as ETAS - to the realm of social media to propose a mutually exciting point process models where each event type (e.g., like) can also trigger sequent events of same or different types (e.g., other likes, but also replies and retweets). Since Ogata's point process casts aftershock events as epidemics, large outbreaks produce much more infections than a small one and, similarly, events triggered by influencers determine a broader cascade. Authors argue the proposed model to be more flexible in accomodating social media dynamics, since it provides a power-law relaxation of the timing activities, and an exponential boost of the responde amplitudes to consider eventual spikes in the content virality. The proposed advancements are supported by an empirical experimentation on a Twitter dataset gathered by the authors. The proposed model - dubbed METAS - has the lowest AIC between all the possible competitors and provides the lowest accuracy error when predicting content popularity. 
\\

In \cite{paper87}, the authors introduce a modelling framework for clustering continuous-time grouped streaming data, the Hierarchical Dirichlet Hawkes process (HDHP), which allows us to automatically uncover a wide variety of learning patterns from detailed traces of learning activity. HDHP is based on a 2-layer hierarchical design. The top layer is a Dirichlet process that determines the learning pattern distribution. The bottom layer corresponds to a collection of independent multivariate Hawkes processes, one per user, with as many dimensions as the number of learning patterns, i.e., infinite. In the HDHP, the popularity of each learning pattern, or equivalently the probability of assigning a new task to it, is constant over time and given by a specific distribution.
\\

The authors in \cite{paper89} address the problem of describing a factorial marked point process for event marker prediction using event history and an event taker's individual level profile. The emphasis is on next-event label estimation, distributed across multiple markers. This work makes two contributions to this problem: A direct solution based on the Alternating Direction Method of Multipliers and the Fast Iterative Shrinkage-Thresholding algorithm, and a reformulation of the original problem into a Logistic Regression model for more efficient learning. In addition, a sparse group regularizer is used to identify the key profile features and event labels. The main differences from previous works are the use of factorial marked point process learning for event marker prediction and a group sparse regularizer for point process learning. The experiments are focused on individual level next job prediction for LinkedIn users and transition prediction to the next ICU department for patients in the MIMIC-II database. The proposed decoupled and reformulated method is tested against state-of-the-art competitors, demonstrating its efficacy.
\\

\subsection{Alternative Approaches}

In \cite{paper31} the authors present a method to predict events related to Online Social Networks (OSNs). The relationships between users of an OSN create \emph{communities}, characterized by a more communicative density in comparison with the rest of the network. However, these relationships change over time and lead to the creation of new communities or the disappearance of others.
The changes occurring in the set of communities over time are called \emph{community events}.
To detect the community events in dynamic OSNs, the authors divide the activity of the network into a sequence of \emph{snapshots}.
In each snapshot a list of communities is created.
To create communities, and detect the events related to them, the authors consider a set of features, divided into (1) qualitative features, based on the degree of centrality, betweenness, closeness, and eigenvector, and (2) qualitative feature as the number of leader nodes of each community.
The events of interests are the \emph{survive}, \emph{spread}, \emph{decompose}, \emph {split} and \emph{die} of the communities.
These events are detected by analyzing the users belonging to the communities.
As an example, a \emph{survive} event occurs for a community at time $t$ if the majority of its users at time $t-1$ belong to it at time $t$, while a \emph{spread} event occurs if the majority of its users at time $t-1$ will move to other communities at time $t$.
\\

In \cite{paper32} the authors propose a method to retrieve the diffusion network given only the cascade events, i.e., the timestamps when each user gets infected. The objective is to identify, for each cascade event, which node was the source triggering the infection. The intuition underlying this work is that the interval between two infections resembles a different distribution if the node pair is connected in the diffusion network w.r.t. when such an edge does not exist. The authors first corroborate this hypothesis through data analysis on a real-world dataset. Then, they define a non-parametric framework that separates the cascades into two clusters: One comprising connected user pairs and another with the rest. The presented clustering framework does not assume any parametric distribution from which the cascade time intervals are sampled. Instead, they use Cramer-Von Mises test statistics to measure the distance between the empirical distributions of the intervals of all pairs of users co-occurring in at least a cascade event. The approach through clustering allows for the mentioned non-parametric formulation since it requires only the distance between cascades. Authors also introduce a novel metric space to speed up the computation of the Cramer-Von Mises test statistic. They show, by means of theoretical analysis, that the algorithm converges to a minimizer after a finite number of steps. The proposed method is further validated through experiments on synthetic and real datasets against several competitors.
\\

Most of the Temporal Point Process methods assume that all event types contribute to the prediction of a target type. However, in reality, some event types cannot give a contribution to predicting the target type, and using them introduce a disturbance in the predicting process.
\cite{paper47} deals with this problem and propose a model that corrects this behavior by learning to discard these event types. The idea is the following. Given an event sequence of $N$ events $H_t=\{(k_i,t_i)\}_{i=1}^N$, where $k_i$ is the event type, $t_i$ is the event timestamp, and $t>t_i$ for each $i \in [1..N]$, an intensity function $\lambda_u(t)=\mu + \sum_{(k_i,t_i)\in H_t} \phi_{u,k}(t-t_i)$, for each event type $u$ is defined. Here the event sequence is assumed to be a Hawkes process where the intensity of the type-$k$ event is increased by the history of the other event types. The function $\phi_{u,k}(\cdot)$ reports whether $k$ is able to influence $u$ or not and is defined as $\phi_{u,k}(t)=G_{u,k}\cdot g(t)$, where $G_{u,k} \in \{0, 1\}$ indicates whether type-$k$ events contribute to predicting type-$u$ events or not, and $g(\cdot)$ is a function modeling the time-decaying influence of a type-$k$ event on the occurrence of a future type-$u$ event.
The matrix $G = [G_{u,k}]$ is the adjacency matrix of a directed graph whose nodes are the event types and edges model the influence between nodes. The proposed model learns $G$ and $g(t)$ simultaneously. In particular, the search for the optimal $G$ is performed using a \emph{random graph method} and the selection of the edges is part of the optimization process. Each node is modeled with an embedding and the probability that an edge between two nodes exists is computed using the related embeddings.
\\

In \cite{paper63} the authors present an interesting unified framework that integrates first-order temporal logic rules as prior knowledge into Point Processes.
Most of the Temporal Point Process methods rely on \emph{deep neural networks} and they are able to model increasingly complex phenomena.
However, they require lots of data to properly fit the models, making them perform poorly in the regime of small data.
Moreover, it is difficult to clearly explain the logic behind their predictions. That is why they have been branded as \emph{black boxes}.
But in many real domains, like in medicine, interpretability is as important as predictions. The goal of the proposed approach is to leverage the prior knowledge from a particular domain - in the form of temporal logic rules - to improve the interpretability of the model. It is particularly suitable in all those cases where the amount of data is small and it is challenging to accurately retrieve these rules via data-driven techniques.
Another advantage in using this model is that the learned logic weights in one dataset can be used to warm-start the learning process on a different dataset. Then, it makes possible to transfer knowledge among different datasets.
The rules used by this framework model temporal relations among events such as “A happens before B”, “if A happens, after 5 mins, B can happen”, and “if A and B happen simultaneously, then at the same time C can happen”.
Each rule is characterized by a conjunction of conditions (\emph{body} of the rule) which, if satisfied, makes a certain event (\emph{head} of the rule) even more likely (or unlikely) to occur.
The \emph{intensity function} of that event is therefore obtained from the linear combination of functions capturing the effects of logic rules on it plus a function modeling the spontaneous occurrence of the event without the influence of the logic component.
\\

The authors in \cite{paper90} propose a novel method named EHTPP for a temporal recommendation. EHTPP fully models user interest, matches data's scale-free distribution, and maintains the model interpretability. EHTPP consists of three major components: In Interaction Embedding, EHTPP considers users, items and continuous time information simultaneously. Specifically, it maps users and items into hyperbolic space while embedding continuous time into low-dimensional vector space. 
In the module of Interaction Generation, EHTPP defines intensity function in the temporal point process to estimate the likelihood of interaction occurring. 
In the Recommendation module, describe the processes of training and recommendation. It is noticeable that the item recommendation is based on the interaction intensity ranking. The greater the intensity, the greater the probability that the item will be recommended. The authors compare the proposed model against several state-of-the-art approaches using three real-world datasets.
\\

\section{Spatio-Temporal Point Processes} \label{sec:stpp}

\begin{table}[]
    \centering
    \resizebox{\linewidth}{!}{
    \begin{tabular}{l|c c c c c c}
         Paper & Data Type & 
         Architecture & 
         Year  \\
         \hline
         \makecell[l]{Deep Mixture Point Processes: \\ Spatio-temporal Event Prediction \\ with Rich Contextual Information} \cite{paper12} &
         \makecell{sets of spatio-temporal points \\ plus related images and text} & 
         \makecell{CNN with attention layers followed \\ by fully connected layers} & 
         2019 \\ 
         \hline
         
         \makecell[l]{Neural Spatio-Temporal Point Processes} \cite{paper36} & 
         Sequence of spatio-temporal events
         &
         \makecell{Neural ODEs with neural jump SDEs \\ and Continuous Normalizing Flows} & 
         2021 \\
         \hline
         
         \makecell[l]{Time Perception Machine: \\ Temporal Point Processes for \\ the When, Where and What of Activity Prediction} \cite{paper59} & 
         Streaming data (videos, trajectories) & 
         \makecell{Hierarchical Recurrent Neural Network \\ with skip connections} & 
         2018\\
         \hline
         
         \makecell[l]{Imitation Learning of Neural Spatio-Temporal \\ Point Processes} \cite{paper61} & 
         Sequence of events (time + location) & 
         
         
         \makecell{Mixture of Gaussian diffusion kernel \\ whose parameters are \\ parameterized by neural networks} & 
         2022 \\
         \hline
         
         \makecell[l]{Interpretable Deep Generative \\ Spatio-Temporal Point Processes} \cite{paper74} & 
         Sequence of spatio-temporal events & 
         
         \makecell{Gaussian Mixture + Neural Network \\ + Imitation Learning}  &
         2020\\
         \hline

         \makecell[l]{Social lstm: Human trajectory prediction \\ in crowded spaces} \cite{paper81} & 
         Sequence of spatio-temporal events &
         LSTM &
         2016\\
         \hline
    \end{tabular}}
    \caption{Spatio-Temporal Point Processes}
    \label{tab:stpp_table}
\end{table}

\subsection{Deep Learning Methods}

\subsubsection{Recurrent Neural Networks}

The goal of \cite{paper59} is to develop an activity forecasting approach to predict the timing, spatial location, and category of the following activities given past information. For this purpose, the authors propose the Time Perception Machine (TPM), a hierarchical recurrent neural network with skip connections for multi-resolution temporal data processing. This approach can learn the temporal distribution of human activities in streaming data (e.g., videos and person trajectories). The input data is sparsely annotated, i.e., only a few semantically meaningful events are in the sequence. The vast majority of works in this field take as input only the annotated frame. The paper's novelty is to use both the annotated frame and the complete sequence, building a hierarchy between the two. This hierarchy is achieved by using two stacked LSTMs, one covering each input frame and one focused on capturing the temporal dynamics of the significant timesteps. The model parameters are learned in a supervised manner by maximizing the likelihood of event sequences. TPM is evaluated on two real-world sports datasets against different baselines: Markov chains for predicting space shift and the category and conventional point processes for time estimation. The experiments empirically demonstrate the proposed methods' efficacy and performance w.r.t. the baselines.
\\

In \cite{paper81}, the authors propose an approach to address both challenges through a novel data-driven architecture for predicting human trajectories in future instants. While LSTMs can learn and reproduce long sequences, they do not capture dependencies between multiple correlated sequences.
The authors address this issue through a novel architecture which connects the LSTMs corresponding to nearby sequences. In particular, they introduce a “Social” pooling layer that allows the LSTMs of spatially proximal sequences to share their hidden states. This architecture, which they refer to as the “Social-LSTM”, can automatically learn typical interactions that occur among trajectories that coincide in time. This model leverages existing human trajectory datasets without needing additional annotations to learn common sense rules and conventions humans observe in social spaces. Two real-world datasets are used to test the effectiveness of the proposed model against state-of-the-art approaches.

\subsubsection{Neural Networks}

In \cite{paper12}, the authors propose DMPP (Deep Mixture Point Processes), a point process model for predicting spatio-temporal events leveraging rich contextual information provided by image and text data (e.g., road networks and social/traffic event descriptions). The goal is to model an intensity function $ \lambda (\cdot)$ that takes as input a spatio-temporal point $\mathbf{x}=(t,s)$, where $t$ is an instant of time and $s$ is a point in a geometric space, and returns the intensity of an event of interest at that point, that is the number of its occurrences around $\mathbf{x}$. The intensity function is designed as a mixture of kernels where the mixture weights are modeled by a deep neural network. In detail, the intensity function is defined as $\lambda(\mathbf{x})=\sum_{j=1}^J f(\mathbf{u}_j, \mathbf{z}_j;\theta) \cdot ker(\mathbf{x}_j,\mathbf{u}_j)$, where $\mathbf{u}_j$ are reference points sampled over the spatio-temporal domain, $\mathbf{z}_j$ are the corresponding contextual information extracted from images or text, $f(\cdot)$ is a deep learning model returning a non-negative scalar, $\theta$ is the set of its parameters, and $ker(\cdot)$ is the kernel function (in the paper the gaussian kernel is used). The number of reference points, $J$, determines the trade-off between accuracy and computation complexity. Learning of the parameters $\theta$ of $f(\cdot)$ can be done with a simple back-propagation process. The deep neural model $f(\cdot)$ used in the paper leverages attention mechanisms that fully exploit visual and textual information. It consists of three components: An image attention network to extract image features, a text attention network to extract text features, and a multimodal fusion module that merges the information provided by the first two modules to derive the intensity output value. The predictive performance of the proposed model is evaluated using three real-world datasets from different domains.
\\

In \cite{paper36}, the authors provide a new class of parameterizations for spatio-temporal point processes. In particular, the authors use the neural ordinary differential equation (ODE) to parameterize the spatio-temporal point process by combining ideas from neural jump SDEs and Continuous Normalizing Flows. For the experimental part, the authors compare the log likelihood obtained from their model with the log-likelihood of several state-of-the-art methods. In particular, the experiments on real datasets show that their approach achieves state-of-the-art performances.
\\

In \cite{paper61}, the authors propose a framework for spatio-temporal point processes and develop an imitation learning-based approach for model fitting. In particular, the authors start by specifying the conditional intensity function of a Hawkes process ($\lambda(t, s|H_t) = \lambda_0 + \sum_{j: t_j < t} v(t, t_j, s, s_j)$) in which they assume that the triggering function (v) that capture the influence of the past events takes the form of a standard Gaussian diffusion kernel over space and decays exponentially over time. In Fig. \ref{fig:paper61} is shown an earthquake example, a typical Hawkes process. They adopt a mixture of Gaussian diffusion kernels whose parameters are parameterized by a deep neural network. To learn parameters for the proposed model, the authors define two approaches: The first is a classical approach based on maximum likelihood estimation, the second is the imitation learning-based approach, a form of reinforcement learning in which a learner takes actions in an environment according to a specific policy and the environment gives feedbacks to the learner via observing the discrepancy between the learner actions and the data provided by the expert. The policy is the probability density function of possible actions given the history and can be expressed with the conditional intensity function. The goal is to find the worst- case reward that maximizes the divergence between the two rewards. The authors also provide a sampling method to generate samples from the proposed model. To evaluate the performance of the proposed model, the authors use the average mean square error and the mean discrepancy metric to show that the data generated are realistic. The experiments show the effectiveness of the proposed model on synthetic and real datasets. In addition, the model is also interpretable by observing the conditional intensity over space at a specific time frame: Its shape is useful for deriving knowledge for domain experts.
\\

In \cite{paper74} the authors present an interesting approach for Spatio-Temporal Point Processes based on Gaussian Mixture. It is called \emph{Neural Embedding Spatio-Temporal (NEST) model}. The idea is to adopt a Spatio-Temporal intensity function obtained from a Gaussian Mixture whose parameters are returned by a neural network trained on the event history.
The Gaussian Mixture is obtained by a set of Gaussian diffusion kernels with shifts, rotations, and non-isotropic shapes.
The authors show that this model is able to capture the complex and heterogenous spatial dependence in discrete events such as earthquakes. The training process is performed using an \emph{imitation learning} (a type of \emph{reinforcement learning}) approach.
\\

\section{Application Areas and Datasets} \label{sec:applications}

In the analyzed papers, several datasets had been used with different settings. In the following, the datasets are categorized based on the application scenario.
For each dataset, we report its description, the papers containing experiments that use it, and, when available, a link to the related web page.

\setlength\rotFPtop{0pt plus 1fil} 
\begin{sidewaystable}
    \centering
    \resizebox{\linewidth}{!}{
    \begin{tabular}{||c|| c | c | c | c ||}
         \textbf{Application Scenario} & \textbf{Datasets} &  \textbf{Description} & \textbf{References} & \textbf{Used in}\\
         \hline
         \hline
          & Stack OverFlow  & \makecell{It is a question-answering website in which users can be awarded \\ by answering posted questions on the website. Each event is composed of the type of reward and \\ the time at which the reward was awarded.} & \cite{StackOverFlow}  & \makecell{\cite{paper3,paper13,paper18,paper20,paper38}\\ \cite{paper37,paper34,paper47,paper73,paper15,paper44}} \\
         \cline{2-5}
         & Retweet & \makecell{It contains sequences of tweets, where each sequence contains \\ an origin tweet and some follow-up tweets. Each tweet has associated the time and \\ the user; users are grouped into categories based on the number of their followers.} & \cite{Retweet} & \makecell{\cite{paper13,paper18,paper20,paper38,paper37} \\ \cite{paper16,paper39,paper47,paper71,paper72}}\\
         \cline{2-5}
         & MemeTrack & \makecell{MemeTrack contains mentions of several memes spanning ten months.} & \cite{MemeTrack} & \makecell{\cite{paper38,paper16,paper39,paper73} \\ \cite{paper88,paper43,paper46}} \\
         \cline{2-5}
         & Twitter(a) & \makecell{Data gathered from Twitter which comprises profiles of 52 million users, 1.9 billion directed follow links \\ among these users, and 1.7 billion public tweets posted by the collected users.} & - & \cite{paper24,paper73}\\
         \cline{2-5}
         & Twitter(b) & \makecell{The dataset contains 569 URLS shared by 5,942 users. The reshare sequence for a URL creates a cascade \\ where the average cascade length is 5.70.} & - & \cite{paper43}\\
         \cline{2-5}
         & LastFM & \makecell{LastFM is a music recommendation dataset containing sequences of songs that selected users listen to over time. \\ Artists are used as event types. Specifically, the dataset contains roughly 1,000 users \\ with entire listening history of 19, 150, 868 (user, time, artist, song) tuples from 2004 to 2009.} & \cite{LastFM} & \cite{paper34,paper41,paper79}\\
         \cline{2-5}
         & Facebook & \makecell{Facebook contains a list of all the posts on the Wall of New Orleans Facebook users. \\ It was used for a period of 23 weeks from March 15, 2007, to August 22, 2007.} & - & \cite{paper31}\\
         \cline{2-5}
         & Reddit & \makecell{Reddit is a social network website in which the users submit posts to subreddits (sub-forums). \\ Posts from the most active users on the most active subreddits are recorded. Each sequence corresponds to \\ a list of submissions a user makes. Subreddits are used as event types.} & \cite{Reddit} & \cite{paper34,paper79,paper80}\\
         \cline{2-5}
         & MOOC & \makecell{MOOC contains the interaction of students with an online course system. There are 97 unique types of interactions.} & - & \cite{paper34}\\
         \cline{2-5}
         & Wikipedia & \makecell{The dataset contains most edited pages. A sequence corresponds to edits of a Wikipedia page.} & - & \cite{paper34} \\
         \cline{2-5}
         & Yelp & \makecell{Yelp contains data from the review forum and the reviews for the 300 most visited restaurants in Toronto are considered. \\ Each restaurant has a corresponding sequence of reviews over time.} & \cite{Yelp} & \cite{paper34,paper80} \\
         \cline{2-5}
         & Duolingo & \makecell{Duolingo contains 13 million Duolingo student learning traces.} & \cite{Duolingo} & \cite{paper24} \\
         \cline{2-5}
         Social media & Microsoft Academic Search (MAS) &  \makecell{MAS provides access to publication venues, time, citations, etc.} & \cite{MAS} & \cite{paper25,paper26}\\
         \cline{2-5}
         & Wiki-Talk & \makecell{The Wikipedia dataset is a time network from Wikipedia in which users edit the discussion page for each other. \\ This dataset has been used for 21 weeks from January 1, 2007, to May 28, 2007, \\ including 104,027 temporary edges during this period.} & \cite{WikiTalk} & \cite{paper31}\\
         \cline{2-5}
         & CollegeMsg & \makecell{The College Massage dataset includes private messages posted on the OSN \\ at the University of California, Irvine. This dataset includes a period of 27 weeks from April 16, 2004, to October 28, 2004, \\ and a total of 59,564 temporary edges. } & \cite{CollegeMsg} & \cite{paper31}\\
         \cline{2-5}
         & Digg & \makecell{Digg is a news aggregator. The dataset contains votes by its users for 3,553 news articles. \\ The timestamp of votes along with the anonymized user-id form the cascade. \\It has 82,778 nodes (users) and an average cascade length of 30.0.} & - & \cite{paper43} \\
         \cline{2-5}
         & Weibo & \makecell{Weibo is a Chinese micro-blogging platform, where the social network \\ consists of follower links, and cascades reflect re-tweeting behavior.} & - & \cite{paper44} \\
         \cline{2-5}
         & Higgs & \makecell{Higgs is a public dataset built by monitoring the spreading processes on \\ Twitter before, during and after the announcement of the discovery of a new particle with \\ the features of the elusive Higgs boson on 4th July 2012.} & - & \cite{paper46} \\
         \cline{2-5}
         & Seismic & \makecell{Seismic data is originally gathered by the authors of \cite{paper14}. \\ It contains a set of Twitter information cascades where each cascade is a set of resharing \\ events and each event is a tuple (u,t) with u being the user out-degree and t being the resharing timestamp.} & \cite{Seismic} & \cite{paper14} \\
         \cline{2-5}
         & \makecell{Amazon Beauty and \\ Amazon Clothes} & \makecell{ Amazon Beauty and Amazon Clothes contain sequence of product reviews.} & - & \cite{paper11}\\
         \cline{2-5}
         & tianchi-mobile & \makecell{tianchi-mobile collects users’ behavioral data on Alibaba’s M-Commerce platforms. \\ It records users’ actions on items with timestamps, rounded to the nearest hour.} & \cite{tianchi_mobile} & \cite{paper41,paper42} \\
         \cline{2-5}
         & ML-1M & \makecell{ML-1M collects users’ rating scores for movies.} & - & \cite{paper90} \\
         \cline{2-5}
         & Behance & \makecell{Behance collects likes and image data from the community art website Behance.} & - & \cite{paper90}\\
         \cline{2-5}
         & Amazon Movies & \makecell{Amazon Movies refers to the reviews of movies. For each item they consider the \\ time of the written review as the time of event in the sequence and the rating (1 to 5) as the corresponding mark.}  & - & \cite{paper73} \\
         \cline{2-5}
         & Amazon Toys & \makecell{Amazon Toys refers to the reviews of toys. For each item they consider the time of the written \\ review as the time of event in the sequence and the rating (1 to 5) as the corresponding mark.} & - & \cite{paper73} \\
         \hline
    \end{tabular}}
    \caption{Dataset Description in Social Media Area}
    \label{tab:dataset_1}
\end{sidewaystable}

\setlength\rotFPtop{0pt plus 1fil} 
\begin{sidewaystable}
    \centering
    \resizebox{\linewidth}{!}{
    \begin{tabular}{||c|| c | c | c | c ||}
         \textbf{Application Scenario} & \textbf{Datasets} &  \textbf{Description} & \textbf{References} & \textbf{Used in}\\
         \hline
         \hline
         & Eletrical Medical Records (MIMIC II) & \makecell{MIMIC II is a collection of de-identified clinical visit records of ICU patients for seven years. Each event records the time when a patient \\ had a visit to the hospital and the disease. The dataset contains 650 sequences, each of which corresponds to a patient’s clinical \\ visits in a seven-year period. Each clinical event records the diagnosis result and the timestamp of that visit.}  & \cite{MIMICII} & \makecell{\cite{paper3,paper13,paper18,paper20} \cite{paper50,paper38,paper37,paper11} } \\
         \cline{2-5}
         & \makecell{Medical Information Mart for \\ Intensive Care III (MIMIC III)} & \makecell{MIMIC-III contains de-identiﬁed clinical visit records from 2001 to 2012 for more than 40,000 patients.} & \cite{MIMICIII} & \cite{paper25,paper63,paper26}\\
         \cline{2-5}
         & CLINIC & \makecell{CLINIC is a dataset for tracking patient clinical status; \\ it consists of baseline physiologic variables along with the time of death (in case).} & \cite{CLINIC} & \cite{paper29}\\
         \cline{2-5}
         & Health & \makecell{Health contains ECG records for patients suffering from heart-related problems.} & - & \cite{paper73} \\
         \cline{2-5}
         Healthcare & FLCHAIN & \makecell{FLCHAIN is a public dataset introduced in a study to determine whether \\ non-clonal serum immunoglobin free light chains are predictive of survival time} & - & \cite{paper69}  \\
         \cline{2-5}
         & SUPPORT & \makecell{SUPPORT is a public dataset introduced in a survival time study of seriously-ill hospitalized adults } & - & \cite{paper69} \\
         \cline{2-5}
         & SEER & \makecell{SEER is a public dataset provided by the Surveillance, Epidemiology, and End Results Program.} & - & \cite{paper69}\\
         \cline{2-5}
         & EHR & \makecell{EHR is a large study from Duke University Health System centered around \\ inpatient visits due to comorbidities in patients with Type-2 diabetes.} & - & \cite{paper69} \\
         \cline{2-5}
         & COVID-19 CASES & \makecell{COVID-19 CASES contains daily COVID-19 cases in New Jersey state. } & - & \cite{paper36} \\
         \cline{2-5}
         & BOLD5000 & \makecell{BOLD5000 consists of fMRI scans as participants are given visual stimuli} & - & \cite{paper36} \\
         \hline
         & Financial Transaction & \makecell{Financial Transaction contains transaction records for a stock in one day. \\ Each record stores the time information (in millisecond) and the possible action, i.e., buy or sell.} & \cite{FinancialTransaction} & \makecell{\cite{paper3,paper50,paper38}\cite{paper11,paper72}} \\
         \cline{2-5}
         & BPI Challenge 2012 & BPI Challenge contains traces of a loan application process at a Dutch financial institute. & \cite{BPI2012} & \cite{paper6,paper9}\\
         \cline{2-5}
         & BPI Challenge 2017 & BPI Challenge contains traces of a loan application process at a Dutch financial institute.  & \cite{BPI_2017} & \cite{paper6,paper9}\\
         \cline{2-5}
         Finance & BPI Challenge 2013 & \makecell{BPI Challenge 2013 contains traces of event data of an incident and problem management system by Volvo Belgium.} & - &\\
         \cline{2-5}
         & NYSE & \makecell{NYSE contains 0.7 million high-frequency trading records from NYSE for a given stock within one day.} & - & \cite{paper25,paper26} \\
         \cline{2-5}
         & FICO-UCSD & \makecell{A credit card dataset from the “UCSD-FICO Data Mining Contest" (FICO-UCSD., 2009) \\ to detect fraud transactions. The dataset is labeled, anonymous and imbalanced.} & - & \cite{paper63}\\
         \cline{2-5}
         & QuantHouse & \makecell{The financial data has been provided by QuantHouse EUROPE/ASIA, and consists of DAX future contracts between 01/01/2014 and 03/01/2014.} & - & \cite{paper88} \\
         \hline
         & Earthquake & \makecell{Earthquake contains the time and location of earthquakes in China.} & - & \cite{paper38}\\
         \cline{2-5}
         Disaster Management & Northern California seismic data & \makecell{The Northern California Earthquake Data Center (NCEDC) provides public \\ time series data that comes from broadband, short period, strong motion seismic sensors, and GPS, and other geophysical sensors.} & - & \cite{paper74,paper61} \\
         \cline{2-5}
         & EARTHQUAKES & \makecell{EARTHQUAKES contains location and time of all earthquakes in Japan from 1990 to 2020 with magnitude of at least 2.5 \\ from the U.S. Geological Survey. Number of events per sequences ranges between 18 to 543.} & - & \cite{paper36} \\
         \hline
    \end{tabular}}
    \caption{Dataset Description in Healthcare, Finance and Disaster Management Areas}
    \label{tab:dataset_2}
\end{sidewaystable}

\setlength\rotFPtop{0pt plus 1fil}
\begin{sidewaystable}
    \centering
    \resizebox{\linewidth}{!}{
    \begin{tabular}{||c|| c | c | c | c ||}
         \textbf{Application Scenario} & \textbf{Datasets} &  \textbf{Description} & \textbf{References} & \textbf{Used in}\\
         \hline
         \hline
         & New York City Taxi & \makecell{New York City Taxi is about the trip records of individual taxis for 12 months in 2013. \\ It contains the temporal information of pick-up (drop-off) passengers associated \\ with every trip as well as the location information.} & \cite{New_York_City_Taxi} & \cite{paper3,paper12,paper73}\\
         \cline{2-5}
         &Breakfast & \makecell{Breakfast dataset is composed of 1712 videos containing 52 subjects performing breakfast preparation activities. \\ The videos are recorded in 18 different kitchens and are composed of 48 fine-grained actions.} & \cite{Breakfast} & \cite{paper1,paper2,paper4,paper70,paper77} \\
         \cline{2-5}
         & 50Salads & \makecell{It contains 50 RGB video and 3-axis accelerometer data belonging to salad preparation activities performed by 25 actors where each actor prepares two salads. \\ There is an annotation file in which the activity is annotated with two levels of granularity (i.e., high and low). \\ Specifically, the dataset is composed of 17 fine-grained action classes. } & \cite{50Salads} & \cite{paper1,paper4,paper70} \\
         \cline{2-5}
         &Multi-THUMOS & \makecell{MultiTHUMOS is a sports activity dataset that is designed for action recognition in videos.\\ The authors derive the CTAS using 400 videos of individuals involved in different sports such as discus throw, baseball, etc. \\ The actions and goals can be classified into 65 and 9 classes respectively and on average, there are 10.5 action class labels per video.} & \cite{Multi_THUMOS} & \cite{paper2,paper77}\\
         \cline{2-5}
         & Helpdesk & \makecell{Helpdesk contains traces from a ticketing management of the help desk of an Italian software company.} & \cite{Helpdesk} & \cite{paper6,paper7,paper9} \\
         \cline{2-5}
         Daily Life & LinkedIn & \makecell{LinkedIn contains the job hopping records of several LinkedIn users \\ in several Information Technologies companies, research institutes and universities.} & \cite{LinkedIn} & \cite{paper50}\\
         \cline{2-5}
         & IPTV & \makecell{IPTV contains log-data records TV watching user behaviors.} & \cite{IPTV} & \cite{paper50,paper80}\\
         \cline{2-5}
         & Smart Home & \makecell{Smart Home contain a sequence from a smart house with 14 classes and over 1000 events. \\ Events correspond to the usage of different appliances. The next event will depend on the time of the day, history of usage and other appliances.} & - & \cite{paper15} \\
         \cline{2-5}
         & Stock & \makecell{Stock contains daily stock prices of 31 companies. We extract three types of event: ‘up’, ‘down’, ’unchanged’ from each stock. \\ We further partition sequences by every season and obtain 1,488 sequences with average length of 49.} & \cite{Stock} & \cite{paper27}\\
         \cline{2-5}
         & eCommerce & \makecell{eCommerce contains users’ behavior in a cosmetics online store. Each user’s behaviors are categorized into four types: \\ ‘view’, ‘cart’, ‘remove-from-cart’, and ‘purchase’. This dataset contains 6,200 sequences with an average length of 64.} & \cite{eCommerce} & \cite{paper27}\\
         \cline{2-5}
         & Citi Bike & \makecell{Citi Bike shares bikes at stations across New York and New Jersey. The activities for a certain bike form a sequence of events. \\ The training set and test set contain the records of the bikes in Jersey City from January to August 2017 and that of September 2017, respectively.} & - & \cite{paper72}\\
         
         \hline
         & 911-Calls & \makecell{911-Calls contains emergency phone call records. The dataset has information about calling time, \\ location of the caller and nature of the emergency.} & \cite{911_Calls} & \cite{paper38,paper25}\\
         \cline{2-5}
         Public Security & Chicago Crime Data & \makecell{Chicago crime data set is a collection of reported incidents of crime that occurred in Chicago; \\ it contains $\sim$ 13 thousand records, each of which shows time, and latitude and longitude of where the crime happened} & - & \cite{paper12}  \\
         \cline{2-5}
         & Atlanta 911 calls-for-services data & \makecell{The 911 calls-for-service data in Atlanta from the end of 2015 to 2017 is provided by the Atlanta Police Department.} & - & \cite{paper74,paper61} \\
         \hline
         & NYC Collision Data & \makecell{New York City vehicle collision (NYC Collision) data set contains $\sim$ 32 thousand motor vehicle collisions. \\ Every collision is recorded in the form of time and location (latitude and longitude coordinates)} & - & \cite{paper12}  \\
         \cline{2-5}
         & \makecell{Global Database of Events, \\ Language, and Tone (GDELT)} & \makecell{GDELT is collected from April 1, 2015 to Mar 31 2016 (temporal granularity of 15 mins). \\ It contains records of events that include two actors, action type and timestamp of event.} & - & \cite{paper67} \\
         \cline{2-5}
         & \makecell{Integrated Crisis Early \\ Warning System (ICEWS)} & \makecell{ICEWS is collected from Jan 1, 2014 to Dec 31, 2014 (temporal granularity of 24 hrs). \\ It contains records of events that include two actors, action type and timestamp of event.} & - & \cite{paper67} \\
         \cline{2-5}
         & VLEP & \makecell{VLEP contains 28,726 examples from 10,234 short video clips.} & \cite{VLEP} & \cite{paper17}\\
         \cline{2-5}
         & Neuron & \makecell{Neuron contains spike record of 219 M1 neurons. Every time a neuron spikes is recorded as an event, \\ these events form a uni-dimensional event sequence. This dataset contains 3,718 sequences with average length of 265.} & \cite{Neuron} & \cite{paper27}\\
         \cline{2-5}
         & Activity-Net & \makecell{Activity-Net comprises of activity categories collected from 591 YouTube \\ videos with a total of 49 action labels and 14 goals.}  & - & \cite{paper77}\\
         \cline{2-5}
         & ETH & \makecell{ETH contains two scenes each with 750 different pedestrians and is split into two sets (ETH and Hotel). } & - & \cite{paper81}\\
         \cline{2-5}
         Others & UCY & \makecell{UCY contains two scenes with 786 people. This dataset has 3-components: ZARA-01, ZARA- 02 and UCY} & - & \cite{paper81} \\
         \cline{2-5}
         & Foursquare & \makecell{An evaluation dataset from Foursquare, a location search and discovery app). \\ Each user has a sequence with the mark corresponding to the type of the check-in location (e.g. "Jazz Club") \\ and the time as the timestamp of the check-in.} & - & \cite{paper73} \\
         \cline{2-5}
         & Celebrity & \makecell{In Celebrity, the authors consider the series of frames extracted from youtube videos of multiple celebrities as event \\ sequences where event-time denotes the video-time and the type is decided upon the coordinates of the frame where the celebrity is located.} & - & \cite{paper73}\\
         \cline{2-5}
         & SMRT & \makecell{SMRT is a set of ATS log data collected from the SMRT corporation, which \\ operates several Mass Rapid Transit systems in Singapore} & - & \cite{paper68} \\
         \cline{2-5}
         & BDD100k & \makecell{BDD100k is a video sequences, accompanied by basic sensory data such as GPS, velocity, or acceleration.} & - & \cite{paper19} \\
         \cline{2-5}
         & NCAA basketball & \makecell{NCAA basketball contains 296 basketball game recordings, each typically 1.5 hours long.} & - & \cite{paper19}\\
         \cline{2-5}
         & GigaWord & \makecell{GigaWord corpus archive of newswire text data in English that has been acquired over several years by the Linguistic Data Consortium}  & - & \cite{paper10}\\
         \hline

         \hline
    \end{tabular}}
    \caption{Dataset Description in Daily Life, Public Security and Others Area}
    \label{tab:dataset_3}
\end{sidewaystable}

\section{Conclusions} \label{sec:conclusions}

In this survey, we investigated the foundations, definitions, methods, and outlooks for event modeling through temporal point processes. We proposed a simplified taxonomy of the approaches and analyzed the main contributions, in order to help readers build a systematic understanding of the area. We also analyzed some relevant application fields and related datasets that can be used for research. 

\section{Acknowledgments}
The authors acknowledge partial support by the EU H2020 ICT48 project “HumanE-AI-Net" under contract \#952026.


\printbibliography[nottype=online, heading=subbibliography, title=REFERENCES]
\printbibliography[type=online, heading=subbibliography, title=DATASET REFERENCES]

\end{document}